\pdfoutput=1

\documentclass[11pt]{article}

\usepackage[]{acl}

\usepackage{times}
\usepackage{latexsym}

\usepackage[T1]{fontenc}

\usepackage[utf8]{inputenc}

\usepackage{microtype}

\usepackage{inconsolata}
\usepackage{amssymb}
\usepackage{algorithm}
\usepackage{graphicx}
\usepackage{multirow}
\usepackage{caption}
\usepackage{subcaption}
\usepackage{soul}
\usepackage{booktabs}
\usepackage{hyperref}
\usepackage{amsmath}
\usepackage{ltablex}
\usepackage{color}
\usepackage{colortbl}

\definecolor{green}{rgb}{0.52, 0.73, 0.4}

\DeclareMathOperator*{\argmax}{argmax}

\title{How are Prompts Different in Terms of Sensitivity?}

\author{
Sheng Lu, Hendrik Schuff, Iryna Gurevych \\ [0.3cm]
Ubiquitous Knowledge Processing Lab (UKP Lab) \\
Department of Computer Science and Hessian Center for AI (hessian.AI) \\
Technical University of Darmstadt \\
\texttt{\small www.ukp.tu-darmstadt.de}
}

\begin{document}
\maketitle
\begin{abstract}

In-context learning (ICL) has become one of the most popular learning paradigms. While there is a growing body of literature focusing on prompt engineering, there is a lack of systematic analysis comparing the effects of prompt techniques across different models and tasks. To address this, we present a comprehensive prompt analysis based on sensitivity. Our analysis reveals that sensitivity is an unsupervised proxy for model performance, as it exhibits a strong negative correlation with accuracy. We use gradient-based saliency scores to empirically demonstrate how different prompts affect the relevance of input tokens to the output, resulting in different levels of sensitivity. Furthermore, we introduce \textit{sensitivity-aware} decoding which incorporates sensitivity estimation as a penalty term in the standard greedy decoding. We show that this approach is particularly helpful when information in the input is scarce. Our work provides a fresh perspective on the analysis of prompts, and contributes to a better understanding of the mechanism of ICL.\footnote{Our code is available at \url{https://github.com/UKPLab/naacl2024-prompt-sensitivity}.}

\end{abstract}

\section{Introduction}

In-context learning (ICL) has become a popular learning paradigm in natural language processing (NLP) due to the rapid development of large language models (LLMs) \cite{brown2020language,dong2022survey,liu2023pre}. With carefully constructed prompts, ICL achieves impressive performance on various tasks \cite{kojima2022large,lampinen-etal-2022-language,wei2022chain,srivastava2022beyond}. As a result, prompt engineering, which aims to find prompts that lead to optimal performance, has emerged as a crucial research topic in ICL \cite{white2023prompt,zhou2022large}. Although effort has been made to understand the effectiveness of certain prompt techniques \cite{feng2023towards,gonen2022demystifying,wang2022towards}, there is no systematic analysis of them across various tasks and models \cite{ajith2023instructeval}. Such an analysis is crucial for prompt engineering, prompt selection, and gaining a deeper understanding of the working mechanism of ICL.


\begin{figure*}
\centering
\begin{subfigure}{0.48\textwidth}
    \centering
    \includegraphics[width=5.6cm]{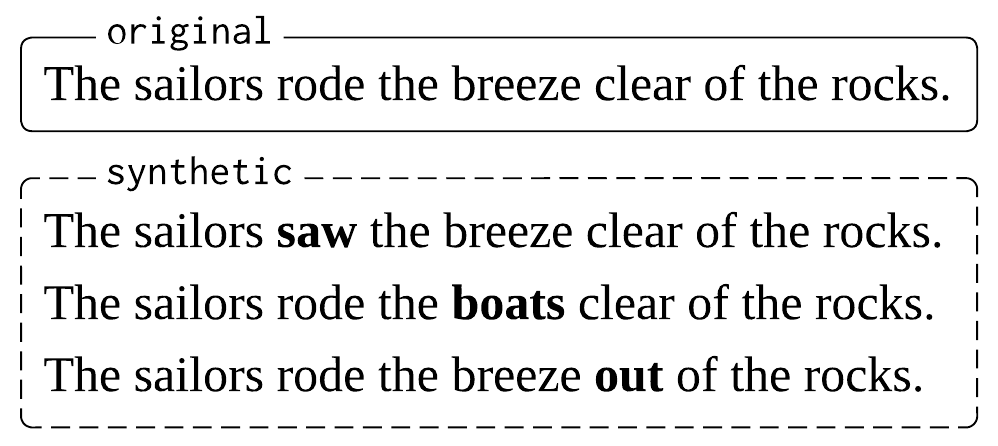}
    \caption{data synthesis}
    \label{diagram_1}
\end{subfigure}
\begin{subfigure}{0.48\textwidth}
    \centering
    \includegraphics[width=5.6cm]{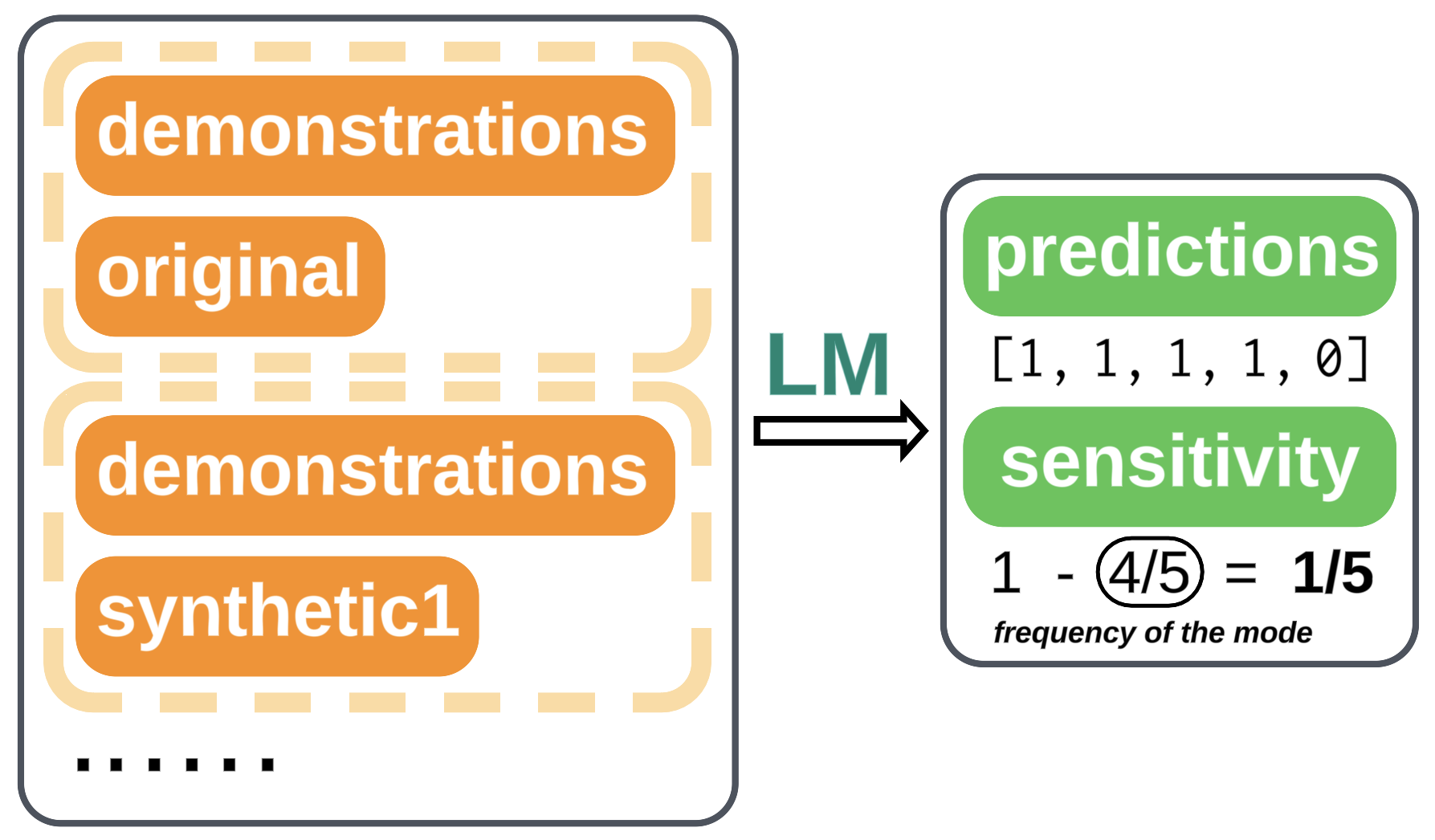}
    \caption{sensitivity estimation}
    \label{diagram_2}
\end{subfigure}
\caption{(a) We generate synthetic data for testing instances using \citet{hahn2021sensitivity}'s framework. (b) We perform inference multiple times using the original and synthetic data, and calculate sensitivity based on the predictions.}
\label{diagram}
\end{figure*}

In this paper, we present a systematic and comprehensive analysis of prompts based on the sensitivity of a function \cite{hahn2021sensitivity}. We hypothesize that certain prompts are more effective for a given task because they decrease the level of sensitivity. Based on the recent findings that ICL implements gradient descent implicitly \cite{akyurek2022learning,li2023transformers,von2023transformers,zhang2023trained}, an effective prompt can be seen as one that facilitates the learning of a new function with lower sensitivity compared to the original function learnt by the model. The sensitivity of a function enables a novel framework to analyze the effect of different prompts. See Figure \ref{diagram} for more details.



We did extensive experiments to validate our hypothesis. We chose five widely used natural language understanding and common sense reasoning tasks. We selected models with varying sizes from three popular families: GPT, LLaMA, and T5. We tested different prompts, including both human-designed prompts and prompts generated by an LLM. The results strongly support our hypothesis that models exhibit different levels of sensitivity depending on the prompts used, and sensitivity is an unsupervised proxy of performance as it has a strong negative correlation between accuracy and sensitivity. With the help of gradient-based saliency scores, we find that tokens in the prompt (e.g., instructions) are more relevant to the output than tokens in the input where perturbations took place, which explains how different prompts lead to varying levels of sensitivity. Furthermore, we introduce \textit{sensitivity-aware} decoding, which incorporates sensitivity estimation as a penalty term in greedy decoding. We show that \textit{sensitivity-aware} decoding is particularly effective when the prompt contains scarce information. Our work provides a fresh perspective for comparing the effects of different prompts and enhances our understanding of the mechanism of ICL.

Our contributions are summarized as follows:
\begin{itemize}
    \item We present a systematic and comprehensive analysis of prompts based on the sensitivity of a function \cite{hahn2021sensitivity}.
    \item We show that sensitivity is an unsupervised proxy of accuracy as it exhibits a strong negative correlation with accuracy.
    \item We use gradient-based saliency scores to show empirically why certain prompts lead to lower sensitivity.
    \item We introduce \textit{sensitivity-aware} decoding and show that it is effective when the prompt contains limited information.
\end{itemize}

\section{Background}

\subsection{In-context learning}

In-context learning (ICL) is a popular learning paradigm that emerged with the advent of LLMs \cite{brown2020language,liu2023pre}. It typically involves prompting the LLM with several demonstrations or exemplars in natural language. Compared to previous learning approaches, ICL has a more interpretable interface and it is more computationally efficient \cite{dong2022survey,zhou2022large}. ICL has demonstrated strong performance on various natural language tasks \cite{kojima2022large,lampinen-etal-2022-language,wei2022emergent,srivastava2022beyond}.

A considerable amount of recent work focuses on revealing the mechanism of ICL. A line of work suggests that ICL is facilitated when the pre-training distribution has certain properties, such as containing compositional structures and latent tasks \cite{chan2022data,hahn2023theory,wies2023learnability}. Empirical evidence shows that ICL implicitly implements gradient descent and constructs a function at inference time \cite{akyurek2022learning,li2023transformers,zhang2023trained}, which may be related to gradient-based meta-learning \cite{von2023transformers}. Similarly, \citet{dai2022can} argue that Transformers are meta-optimizers which produce meta-gradients according to the demonstrations through forward pass, and these meta-gradients are applied to the model through attention.


\subsection{Prompt engineering}

Prompt engineering is essential to effectively retrieving information from an LLM \cite{reynolds2021prompt,schick2021few,white2023prompt,zhou2022large}. An LLM usually requires careful prompt engineering, since a model may not understand prompts in the way a human does \cite{webson2021prompt}.

In this work, we focus on discrete prompts, i.e., prompts that are described in natural language phrases \cite{liu2023pre}. We follow \citet{dong2022survey} and categorize discrete prompts into human-designed and LM-generated prompts, depending on whether they are written by humans or generated by a language model (LM).




\subsection{Prompt analysis}

Most of the existing analytical work concentrates on understanding a particular type of prompt \cite{feng2023towards,gonen2022demystifying,wang2022towards}. However, there is a lack of systematic analysis that compares the effects of different prompts across various models and tasks. As far as we are aware, \citet{ajith2023instructeval} is the only work that presents a systematic analysis of prompts. They evaluate the effect of popular instruction selection methods, whereas our work examines a broader range of prompts.

\color{blue}
\color{black}

\subsection{Sensitivity}
\label{sensitivity}

Previous studies have primarily focused on analyzing sensitivity at the instance level. These investigations reveal that ICL performance is highly dependent on demonstrations, such as the selection of exemplars and the order in which they are presented \cite{zhao2021calibrate,liu2021makes,lu2021fantastically,ajith2023instructeval,chang2022careful,chen2022relation,wang2023large,sclar2024quantifying}. Moreover, \citet{chen2022relation} observe that predictions sensitive to perturbations are more likely to be incorrect.

Based on the theory of Boolean function sensitivity, \citet{hahn2021sensitivity} propose sensitivity as a theory of complexity for sequence classification tasks. The sensitivity of a function quantifies the number of disjoint subsets of the input sequence that can be changed in such a way as to change the output. In the setting of sequence classification, sensitivity measures the non-linearity of the decision boundary. Low-sensitivity tasks are those where low-sensitivity functions, such as linear classifiers, are most successful. High-sensitivity tasks, on the other hand, require high-sensitivity methods, which are more complex. The amount of information in the input is a key factor of sensitivity. Intuitively, if a single change in the input completely changes the output, it is believed that the input does not contain sufficient information, resulting in high sensitivity. An output is more stable if there is redundant information in the input, which is an indicator of low sensitivity.

Sensitivity is an indicator of both architectural and task complexity, and thus it is used as a hardness measure in many NLP tasks \cite{richardson2022pushing,zhao2022measuring,bhattamishra-etal-2023-simplicity}.


\section{Experiment settings}

We use \citet{hahn2021sensitivity}'s framework to generate perturbed data.\footnote{See \url{https://github.com/m-hahn/sensitivity}.} Each of the synthetic data agrees on the original instance on all indices outside a subset. We notice that the synthetic data for one particular dataset are noisier (see Table \ref{noisy_data_csqa} in \ref{more_on_experiment_settings} for a manual inspection of the data). This does not pose an issue because this dataset is our control variable.



\noindent\textbf{Sensitivity estimation}\  The sensitivity estimation proposed in \citet{hahn2021sensitivity} uses the variance of the outputs. We adopt a more straightforward alternative, variation-ratio \cite{freeman1965elementary}, to estimate sensitivity. Given an original input and $n$ synthetic inputs, sensitivity is calculated as
\begin{equation}
    s = 1 - \frac{f_m}{n + 1},
\end{equation}
\noindent where $f_m$ is the frequency of the mode of the $n + 1$ predictions, i.e., the prediction for the original input plus $n$ predictions for the synthetic inputs. The lower $s$ is, the less sensitive a model is to an input.


\noindent\textbf{Dataset}\  We picked five commonly used natural language understanding and reasoning tasks: CoLA \cite{warstadt2019neural}, MultiNLI \cite{williams2017broad}, RTE \cite{wang2019superglue}, SST2 \cite{socher2013recursive}, and CSQA \cite{talmor2018commonsenseqa}. We experimented with GSM8K \cite{cobbe2021gsm8k}, a collection of arithmetic reasoning problems, to assess sensitivity in open-ended generation.

\noindent\textbf{Model}\  We tested models with different architectures and sizes selected models from three popular model families: OpenAI \texttt{text-davinci-003} (GPT3.5-175B), GPT-JT-6B \cite{gpt-j}, LLaMA2-13B-chat, LLaMA2-7B-chat \cite{touvron2023llama2}, Flan-T5-11B, and Flan-T5-770M \cite{chung2022scaling}.

\noindent\textbf{Prompt}\  Table \ref{prompt_reference} shows the prompts we used in our experiments. We experimented with both human-designed and LM-generated prompts. We designed \texttt{base\_a} and \texttt{base\_b} as two baseline prompts. Compared to the simplest \texttt{base\_a} which contains plain \textit{input-target} pair, \texttt{base\_b} includes a human-designed instruction. We designed \texttt{zero\_a} and \texttt{zero\_b} to test an extreme case, where the ground truth is included in the prompt. We tested two popular prompts, i.e., context faithful prompting (\texttt{CFP}) \cite{zhou2023context} and \textit{Chain-of-Thought} prompting (\texttt{CoT}) \cite{wei2022chain}.\footnote{\texttt{CoT} prompting was only tested with the larger models, i.e., GPT3.5-175B, Flan-T5-11B, and LLaMA2-13B-chat.} For LM-generated prompts, we tested automatic prompt engineer (\texttt{APE}) \cite{zhou2022large} and generated knowledge prompting (\texttt{GKP}) \cite{liu2021generated}.\footnote{\texttt{APE} was only tested on CoLA and RTE using GPT3.5-175B due to budget constraints. \texttt{GKP} was only tested on CSQA, and we used the knowledge generated by \citet{liu2021generated}.} In addition, we map each option to an index, so that we can better control the format of the output.

See \ref{more_on_experiment_settings} for more details regarding the setup of experiments.

\begin{table}[h]
\small
\centering
\begin{tabular}{cp{0.76\linewidth}}
\toprule
\textbf{prompt} & \textbf{text} \\ \midrule
\multirow{2}{*}{\texttt{base\_a}} & I'm glad I saw anybody. \\
& \texttt{\{target\}} \\ \midrule
\multirow{4}{*}{\texttt{base\_b}} & \textbf{\textsc{sentence:}} I'm glad I saw anybody. \\
& \textbf{\textsc{question:} Is this (0) unacceptable, or (1) acceptable?} \\
& \textbf{\textsc{answer:}} \texttt{\{target\}} \\ \midrule
\multirow{2}{*}{\texttt{zero\_a}} & I'm glad I saw anybody. \textbf{The answer is 0.} \\
& \texttt{\{target\}} \\ \midrule
\multirow{3}{*}{\texttt{zero\_b}} & \textsc{sentence:} I'm glad I saw anybody. \textbf{The answer is 0.} \\
& \textsc{answer:} \texttt{\{target\}} \\ \midrule
\multirow{4}{*}{\texttt{CFP}} & \textbf{Bob said,} ``I'm glad I saw anybody.'' \\
& \textsc{question:} Is this (0) unacceptable, or (1) acceptable \textbf{in Bob's opinion}? \\
& \textsc{answer:} \texttt{\{target\}} \\ \midrule
\multirow{7}{*}{\texttt{CoT}} & \textsc{sentence:} I'm glad I saw anybody. \\
& \textsc{question:} Is this (0) unacceptable, or (1) acceptable? \\
& \textsc{answer:} \textbf{Let's think step by step.} This sentence is ungrammatical because ``anybody'' is used as the object in an affirmative clause. So the answer is \texttt{\{target\}} \\ \midrule
\multirow{5}{*}{\texttt{APE}} & \textbf{\textsc{instruction:} determine whether each sentence was (1) acceptable or (0) unacceptable based on its structure and grammar.} \\
& \textsc{input:} I'm glad I saw anybody. \\
& \textsc{output:} \texttt{\{target\}} \\ \midrule
\multirow{7}{*}{\texttt{GKP}} & \textbf{\textsc{knowledge:} Electronic maps are the modern version of paper atlas.} \\
& \textsc{input:} Google Maps and other highway and street GPS services have replaced what? \\
& \textsc{options:} (0) united states, (1) mexico, (2) countryside, (3) atlas, (4) oceans \\
& \textsc{output:} \texttt{\{target\}} \\
\bottomrule
\end{tabular}
\caption{Examples of prompts used in our experiments. Contents that are characteristic to a prompt are \textbf{bolded}.}
\label{prompt_reference}
\end{table}

\section{Results}
\label{results}

\begin{figure}[h]
\centering
\includegraphics[width=6.8cm]{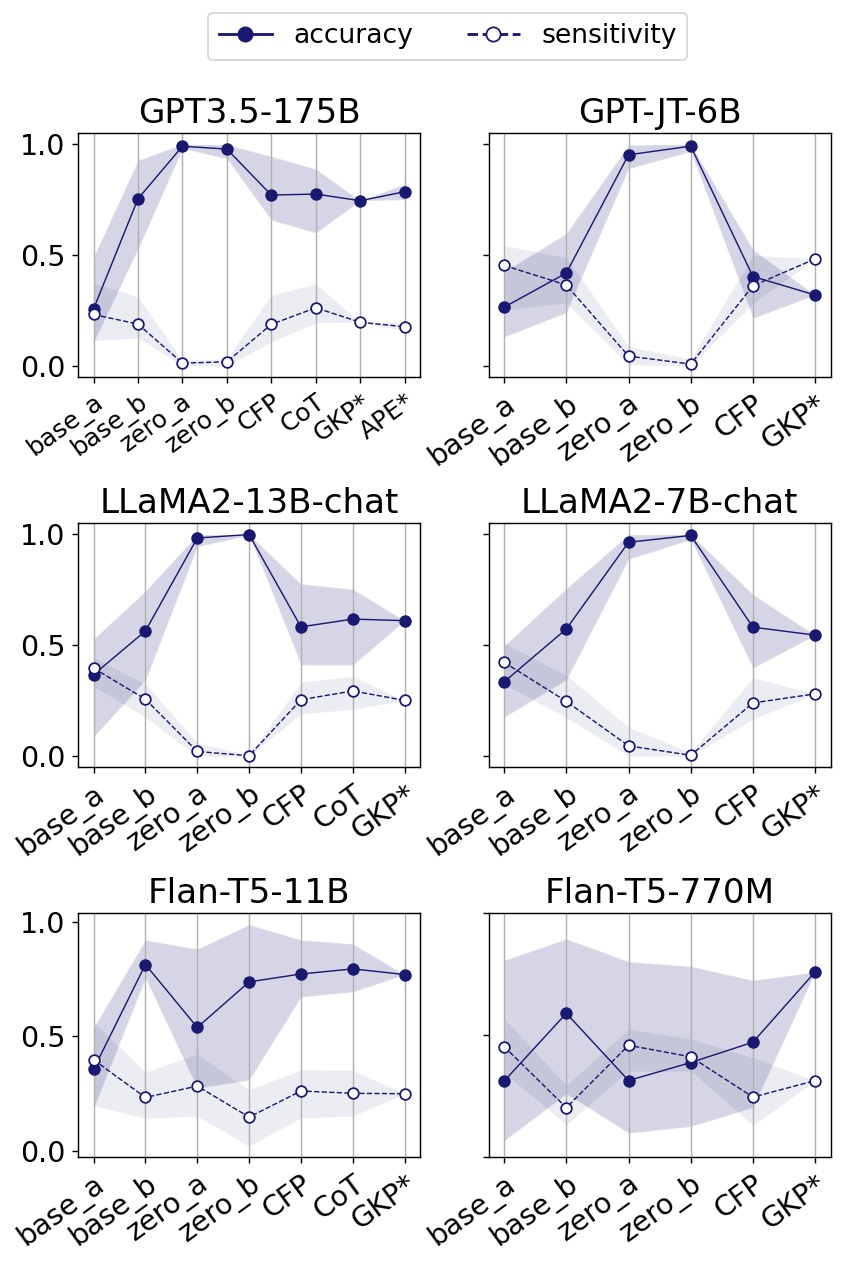}
\caption{The average accuracy and sensitivity of each model using various prompts across different datasets. \texttt{*} indicates prompts that are not tested on all datasets.}
\label{accuracy_sensitivity_correlation}
\end{figure}

Figure \ref{accuracy_sensitivity_correlation} shows the average accuracy and sensitivity of each model using various prompts across different datasets.\footnote{Due to limited space, we did not include the standard deviations of the statistics in the following plots and tables. For more results, please refer to \url{https://github.com/UKPLab/naacl2024-prompt-sensitivity}.} The \textit{Pearson} correlation coefficient shows a strong negative correlation between accuracy and sensitivity ($r=-0.8764$, \textit{p}-value $\ll0.01$). Sensitivity can be viewed as an unsupervised proxy of accuracy given such a strong correlation.




It is interesting to note that both Flan-T5 models failed with \texttt{zero\_a} and \texttt{zero\_b}. We discuss this further in Section \ref{why_flan_t5_failed_with_zero}.

\subsection{Instruction, knowledge, \textit{chain-of-thought}}

This section compares the effects of instruction (human-designed instructions in \texttt{base\_b} and LM-generated instructions in \texttt{APE}), reasoning chain (in \texttt{CoT}), and knowledge (in \texttt{GKP}).



Table \ref{base_b_ape} compares the performance of models using \texttt{base\_b} and \texttt{APE}. The two prompts lead to similar accuracy and sensitivity on CoLA and RTE, suggesting that human-designed and LM-generated instructions have similar effects on the model. At least for CoLA and RTE, there is no need to generate instructions using an LM.

\begin{table}[h]
\small
\centering
\begin{tabular}{lccc}
\toprule
\textbf{dataset}      & \textbf{prompt}       & \textbf{accuracy$\uparrow$} & \textbf{sensitivity$\downarrow$}     \\ \midrule
\multirow{2}{*}{CoLA} & \texttt{base\_b}      & \textbf{0.8235} & \textbf{0.1830} \\
                      & \texttt{APE}          & 0.8216          & 0.1960          \\ \midrule
\multirow{2}{*}{RTE}  & \texttt{base\_b}      & 0.6931          & \textbf{0.1377} \\
                      & \texttt{APE}          & \textbf{0.7509} & 0.1603          \\
\bottomrule
\end{tabular}
\caption{The accuracy and sensitivity of GPT3.5-175B using \texttt{base\_b} and \texttt{APE}.}
\label{base_b_ape}
\end{table}

Table \ref{base_b_gkp} shows that \texttt{GKP} leads to a higher accuracy and lower sensitivity in most cases. This suggests that the effects of instructions and LM-generated knowledge are cumulative, i.e., placing LM-generated knowledge before instructions yields better results.

\begin{table}[t]
\small
\centering
\begin{tabular}{lcccc}
\toprule
\multirow{2}{*}{\textbf{model}} & \multicolumn{2}{c}{\textbf{accuracy$\uparrow$}} & \multicolumn{2}{c}{\textbf{sensitivity$\downarrow$}} \\ \cmidrule{2-5}
                       & \texttt{base\_b} & \texttt{GKP} & \texttt{base\_b} & \texttt{GKP}                    \\ \midrule
GPT3.5-175B              & \textbf{0.8000}        & 0.7459          & 0.3133          & \textbf{0.1982}         \\
GPT-JT-6B              & 0.2413        & \textbf{0.3202}          & 0.4912          & \textbf{0.4831}         \\
LLaMA2-13B             & 0.6085        & \textbf{0.6109}          & 0.3257          & \textbf{0.2511}         \\
LLaMA2-7B              & 0.5276        & \textbf{0.5456}          & 0.3649          & \textbf{0.2804}         \\
Flan-T5-11B            & \textbf{0.8057}        & 0.7697          & 0.3435          & \textbf{0.2491}         \\
Flan-T5-770M           & 0.2607        & \textbf{0.7601}          & \textbf{0.2345}          & 0.3141         \\ \midrule
\textsc{average}       & 0.5082        & \textbf{0.6103}          & 0.3485          & \textbf{0.3082}         \\
\bottomrule
\end{tabular}
\caption{The accuracy and sensitivity using \texttt{base\_b} and \texttt{GKP} on CSQA. LLaMA2-13B and LLaMA2-7B are LLaMA2-13B-chat and LLaMA2-7B-chat.}
\label{base_b_gkp}
\end{table}

Table \ref{CoT_and_task_info_1} shows that \texttt{CoT} leads to a similar accuracy but higher sensitivity compared to \texttt{base\_b}. As shown in Table \ref{prompt_reference}, instructions are also contained in \texttt{CoT}. Unlike LM-generated knowledge, reasoning chains do not bring performance gain on top of instructions.

\begin{table}[ht]
\small
\centering
\begin{tabular}{lcccc}
\toprule
\multirow{2}{*}{\textbf{model}} & \multicolumn{2}{c}{\textbf{accuracy$\uparrow$}} & \multicolumn{2}{c}{\textbf{sensitivity$\downarrow$}} \\ \cmidrule{2-5}
                 & \texttt{base\_b} & \texttt{CoT} & \texttt{base\_b} & \texttt{CoT}                    \\ \midrule
GPT3.5-175B        & 0.7549 & \textbf{0.7758} & \textbf{0.1899} & 0.2636 \\
LLaMA2-13B       & 0.5642 & \textbf{0.6179} & \textbf{0.2564} & 0.2939 \\
Flan-T5-11B      & \textbf{0.8134} & 0.7943 & \textbf{0.2328} & 0.2509 \\ \midrule
\textsc{average} & 0.7108          & \textbf{0.7293}          & 0.2264          & \textbf{0.2695}         \\
\bottomrule
\end{tabular}
\caption{The accuracy and sensitivity of GPT3.5-175B, LLaMA2-13B-chat (LLaMA2-13B), and Flan-T5-11B using \texttt{base\_b} and \texttt{CoT} across different tasks.}
\label{CoT_and_task_info_1}
\end{table}

We designed \texttt{CoT\_base\_a} to isolate the effect of reasoning chains. \texttt{CoT\_base\_a} is a combination of \textit{chain-of-thought} and the most basic \texttt{base\_a} (see Table \ref{prompt_reference_cot_standard_a} in \ref{more_on_instruction_chain_of_thought_and_knowledge}).

\begin{figure}[h]
\centering
\includegraphics[width=7.6cm]{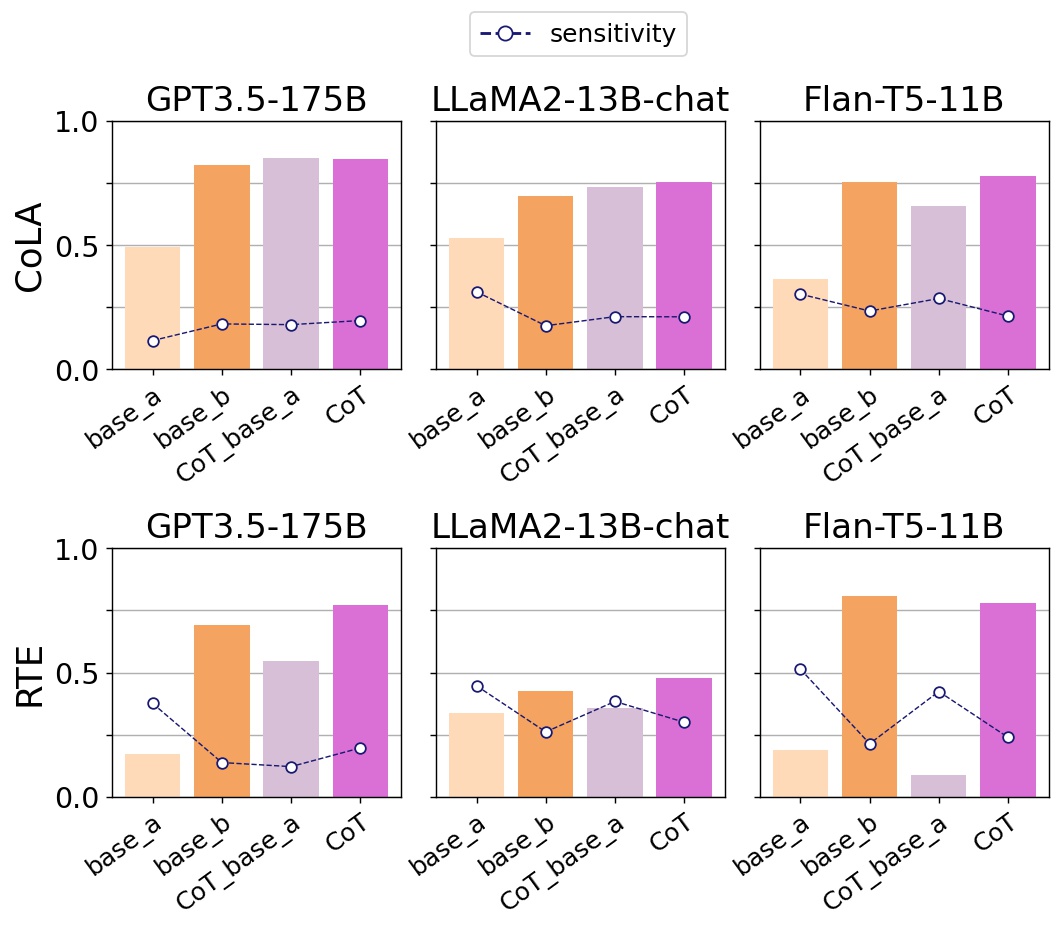}
\caption{The accuracy and sensitivity of different models using \texttt{base\_a}, \texttt{base\_b}, \texttt{CoT\_base\_a}, and \texttt{CoT}.}
\label{CoT_and_task_info_2}
\end{figure}

Figure \ref{CoT_and_task_info_2} shows that \texttt{CoT\_base\_a} outperforms \texttt{base\_a}, but it performs worse than \texttt{base\_b} in most cases. This suggests that for CoLA and RTE, while reasoning chains do help improve performance, it is not as effective as instructions.

We note that models using \texttt{CoT\_base\_a} perform relatively better on CoLA than on RTE. We speculate that this is because the demonstrations in \texttt{CoT\_base\_a} for CoLA contain an equivalent of an instruction. Since CoLA is a binary classification task with \textsc{grammatical} and \textsc{ungrammatical} being the labels, a reasoning chain, such as the one exemplified in Table \ref{prompt_reference_cot_standard_a}, contain an explicit label mapping, which may function similarly to an instruction.

\subsection{What happened to Flan-T5 with \texttt{zero}?}
\label{why_flan_t5_failed_with_zero}

As shown in Figure \ref{accuracy_sensitivity_correlation}, unlike other models, both Flan-T5-11B and Flan-T5-770M failed with \texttt{zero\_a} and \texttt{zero\_b}. We examine the outputs of Flan-T5-11B closely to investigate this counter-intuitive phenomenon, and find that Flan-T5-11B tends to produce text answers instead of numeric indices with \texttt{zero\_a} and \texttt{zero\_b} (see Figure \ref{flan_t5_with_zero} in \ref{more_on_flan_t5_with_zero}). Our observations suggest that Flan-T5 models are not good at mapping the numeric indices in the examplars to their output spaces, unless explicitly instructed via instructions such as in \texttt{base\_b} and \texttt{APE}, or in the form of \textsc{options}, such as in \texttt{GKP}. This conclusion is also supported by the observation that Flan-T5 models fail to produce numeric indices with \texttt{base\_a} (see Table \ref{t5_with_base_a} in \ref{more_on_flan_t5_with_zero}), where explicit instructions are also lacking.

\subsection{The effect of decoding strategies}

\begin{figure}[h]
\centering
\includegraphics[width=6.6cm]{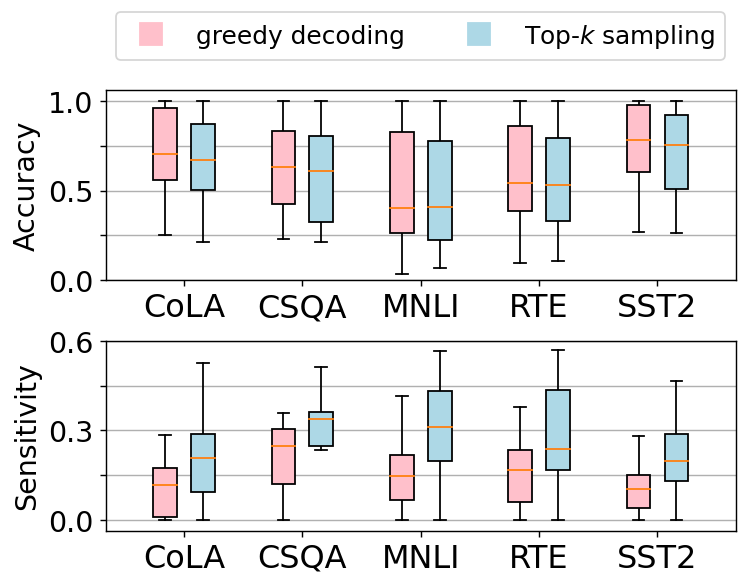}
\caption{The accuracy and sensitivity of predictions obtained using greedy decoding and Top-\textit{k} sampling across different models.}
\label{greedy_top_k}
\end{figure}

It has been shown that decoding strategies influence the quality of LLM generations \cite{lee2022factuality,wang2023selfconsistency}. We also observe that decoding strategies have an effect on sensitivity. Figure \ref{greedy_top_k} shows that the overall sensitivity calculated from predictions obtained using greedy decoding is lower than that of those obtained using Top-\textit{k} sampling \cite{fan-etal-2018-hierarchical}. A strong negative correlation between accuracy and sensitivity is still observed in this case ($r=-0.5507$, \textit{p}-value $\ll0.01$). This is lower than that of Top-\textit{k} sampling ($r=-0.8764$, \textit{p}-value $\ll0.01$), suggesting that Top-\textit{k} sampling has a ``magnifying'' effect on sensitivity. 


In Top-\textit{k} sampling, the next token is sampled from a probability mass redistributed among tokens that have the highest \textit{k} probabilities. The observation that Top-\textit{k} sampling leads to a higher level of sensitivity indicates that for instances with high sensitivity, the output probabilities of different labels are close.

\subsection{Open-ended generation}

Measuring (or even defining) sensitivity in open-ended generation can be challenging. Two pieces of generated text can convey the same meaning even if they vary significantly in terms of word choices and length. We circumvent this issue by selecting an open-ended generation task where the output is more ``controllable.'' Specifically, we chose GSM8K, an arithmetic reasoning task in which the outputs are numbers \cite{cobbe2021gsm8k}. Despite being an open-ended generation task, the numerical format of the outputs in GSM8K allows us to use variation-ratio to measure sensitivity. Our results reveal that there is also a negative correlation between accuracy and sensitivity in the open-ended setting. See \ref{more_on_open_ended_generation} for more details.

\section{Gradient-based saliency scores}
\label{gradient_base_saliency_scores}

In light of recent studies that link ICL and implicit gradient descent \cite{akyurek2022learning,li2023transformers,von2023transformers,zhang2023trained}, we investigate the relationship between sensitivity and gradient. We use gradient-based saliency scores, which reveal the relevance of input tokens to the model prediction \cite{simonyan2013deep,li-etal-2016-visualizing,yin2022interpreting}.
The higher the score is, the more a token is supposed to contribute to the model output. 
We compute gradient-based saliency scores based on the norm of the gradient of the model output.
The gradient $g$ for a token $x_i$ in an input $\boldsymbol{x}$ is calculated as follows:
\begin{equation}
    g(x_i) = \nabla_{x_i}M(y|\boldsymbol{x}),
\end{equation}
\noindent where $M(y|\boldsymbol{x})$ is the logit for the output token $y$. The saliency score $S(x_i)$ is obtained by taking the L1 norm of $g(x_i)$:
\begin{equation}
    S(x_i) = \| g(x_i) \|_{L1}.
\end{equation}

\begin{figure*}[ht]
\centering
    \includegraphics[width=15.6cm]{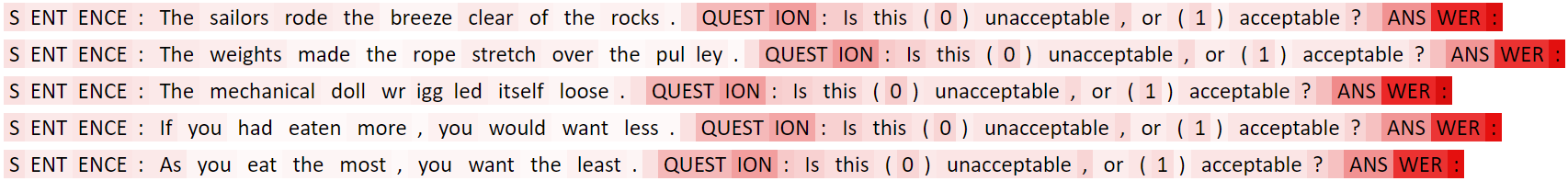}
\caption{Saliency scores over tokens of CoLA instances with \texttt{base\_b} obtained using GPT-6B-JT.}
\label{saliency_base_b}
\end{figure*}

We calculate the saliency scores using GPT-JT-6B, GPT-J-6B, Flan-T5-770M, and T5-770M. Figure \ref{saliency_base_b} gives an example of saliency scores over tokens, which shows that the tokens in the prompt are more relevant to the output than tokens in the input (see Figure \ref{saliency} in \ref{more_on_gradient_based_saliency_scores} for more examples).

\begin{figure}[H]
\centering
\includegraphics[width=6cm]{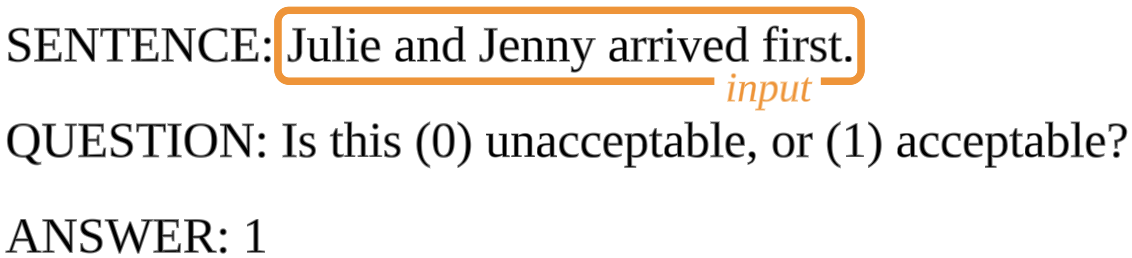}
\caption{An example of token segmentation for instances with \texttt{base\_b}.}
\label{segmentation_base_b}
\end{figure}

To perform a quantitative analysis, we segment an instance into several parts. Take Figure \ref{segmentation_base_b} as an example.

\noindent For a sentence $X$, tokens where perturbations happen are referred to as $X_{input}$ or \textit{input} tokens (such as the \textit{input} in Figure \ref{segmentation_base_b}), and the rest of the sentence is referred to as $X_{prompt}$ or \textit{prompt} tokens.\footnote{For the segmentation of other prompts, please refer to Figure \ref{more_on_segmentation} in \ref{more_on_gradient_based_saliency_scores}.} We calculate the \textit{mean saliency score}, denoted as $\overline{S}$, for input tokens and prompt tokens respectively:
\begin{equation}
    \overline{S} = \frac{\sum_{x_i \in X} S(x_i)}{n}, X \in \{X_{input}, X_{prompt}\},
\end{equation}
\noindent where $n$ is the number of tokens in $X_{input}$ or $X_{prompt}$.

Table \ref{mean_saliency_scores} shows the average \textit{mean saliency scores} of instances in different datasets. Similar to Figure \ref{saliency_base_b}, the \textit{mean saliency scores} of \textit{input} tokens are consistently lower than those for \textit{prompt} tokens. There is a strong negative correlation between $\overline{S}_p - \overline{S}_i$ and sensitivity ($r=-0.7596$, \textit{p}-value $\ll0.01$).\footnote{The \textit{Pearson} correlation coefficient for $\overline{S}_p - \overline{S}_i$ and sensitivity without \texttt{zero\_b} results is $-0.5733$ (\textit{p} $=0.0831$).} This explains why \texttt{base\_b}, \texttt{zero\_b}, and \texttt{CFP} lead to lower levels of sensitivity: perturbations are only done to \textit{input} tokens, which are less relevant to the outputs than \textit{prompt} tokens. We also find that instruction tuned models ``focus'' more on \textit{prompt} tokens than their non-instruction tuned counterparts (see Table \ref{mean_saliency_scores_gpt_jt_gpt_j} in \ref{more_on_gradient_based_saliency_scores}).

\begin{table}[ht]
\small
\centering
\begin{tabular}{lccccc}
\toprule
\multicolumn{1}{l}{\textbf{dataset}} & \multicolumn{1}{c}{\textbf{prompt}}          & $\overline{S}_i$\  & $\overline{S}_p$  & $\overline{S}_p - \overline{S}_i$ & \textbf{sens.$\downarrow$} \\ \midrule
\multirow{3}{*}{CoLA}      & \texttt{base\_b}    & 4.17          & 12.74         & 8.57                 & 30.15          \\
                           & \texttt{zero\_b}    & 7.65          & 23.66         & 16.01                & 0.00           \\
                           & \texttt{CFP}        & 3.75          & 11.43         & 7.68                 & 28.39          \\ \midrule
\multirow{3}{*}{CSQA}      & \texttt{base\_b}    & 2.68          & 7.41          & 4.73                 & 48.84          \\
                           & \texttt{zero\_b}    & 2.25          & 8.02          & 5.77                 & 0.00           \\
                           & \texttt{CFP}        & 2.37          & 7.06          & 4.69                 & 49.94          \\ \midrule
\multirow{3}{*}{MNLI}      & \texttt{base\_b}    & 3.10          & 12.79         & 9.69                 & 43.78          \\
                           & \texttt{zero\_b}    & 3.61          & 17.18         & 13.57                & 3.06           \\
                           & \texttt{CFP}        & 1.68          & 6.71          & 5.03                 & 42.70          \\ \midrule
\multirow{3}{*}{RTE}       & \texttt{base\_b}    & 1.95          & 9.48          & 7.53                 & 30.61          \\
                           & \texttt{zero\_b}    & 3.22          & 18.95         & 15.73                & 0.36           \\
                           & \texttt{CFP}        & 1.53          & 7.32          & 5.79                 & 29.82          \\ \midrule
\multirow{3}{*}{SST2}      & \texttt{base\_b}    & 3.09          & 12.59         & 9.50                 & 28.17          \\
                           & \texttt{zero\_b}    & 4.99          & 22.32         & 17.33                & 0.25           \\
                           & \texttt{CFP}        & 2.77          & 10.70         & 7.93                 & 28.83          \\
\bottomrule
\end{tabular}
\caption{The average \textit{mean saliency scores} of \textit{input} tokens ($\overline{S}_i$, in permillage), \textit{prompt} tokens ($\overline{S}_p$, in permillage) of instances, the difference between the two scores ($\Delta$), and sensitivity (\textbf{sens.}, in percentage) obtained using GPT-JT-6B.}
\label{mean_saliency_scores}
\end{table}

\begin{figure}[H]
\centering
\includegraphics[width=7.6cm]{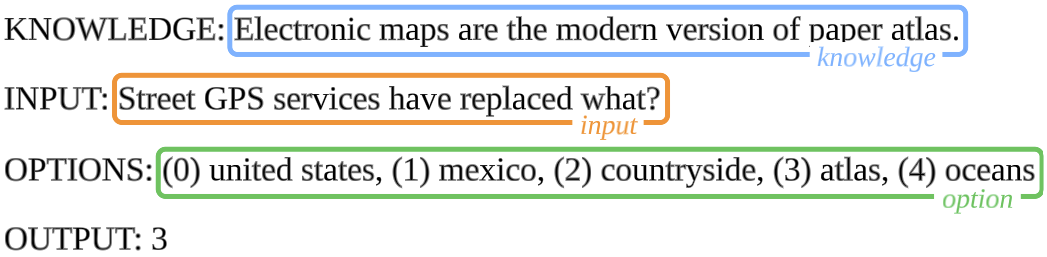}
\caption{An example of token segmentation for instances with \texttt{GKP}.}
\label{segmentation_gkp}
\end{figure}

For \texttt{GKP}, we perform a more detailed segmentation for a better analysis, which is shown in Figure \ref{segmentation_gkp}. We segment an instance with \texttt{GKP} to \textit{knowledge}, \textit{input}, \textit{option}, and \textit{prompt} tokens.

Table \ref{mean_saliency_scores_gkp} shows the average \textit{mean saliency scores} of tokens in CSQA instances using \texttt{GKP}. Similar to Table \ref{mean_saliency_scores}, \textit{input} tokens are less relevant to the predictions. Note that \textit{knowledge} tokens have the lowest average \textit{mean saliency scores}, suggesting that generated knowledge is not very relevant to the predictions.

\begin{table}[h]
\small
\centering
\begin{tabular}{cccc}
\toprule
$\overline{S}_{input}$ & $\overline{S}_{knowledge}$ & $\overline{S}_{option}$ & $\overline{S}_{prompt}$ \\ \midrule
4.33 & 2.56 & 6.37 & 12.86 \\
\bottomrule
\end{tabular}
\caption{The average \textit{mean saliency scores} ($\overline{S}$) of \textit{input}, \textit{knowledge}, \textit{option}, and \textit{prompt} tokens of CSQA instances with \texttt{GKP} obtained using GPT-JT-6B.}
\label{mean_saliency_scores_gkp}
\end{table}

The percentage of $\overline{S}_{input}$ for \texttt{base\_b} (27.2\%) is higher than that for \texttt{GKP} (16.6\%), which indicates that \textit{input} tokens in \texttt{base\_b} are relatively more relevant to the predictions than those in \texttt{GKP}. This is consistent with the observation in Table \ref{base_b_gkp}, that \texttt{GKP} leads to a lower level of sensitivity in most cases.

We also examine the \textit{mean saliency scores} of ground truth tokens in \texttt{zero\_b} (see Table \ref{mean_saliency_scores_for_target_tokens} in \ref{more_on_gradient_based_saliency_scores}). The results show that Flan-T5-770M ``focuses'' less on ground truth tokens than GPT-JT-6B, which explains the failure of Flan-T5 models with \texttt{zero} prompts discussed in Section \ref{why_flan_t5_failed_with_zero}.

\begin{table*}[h]
\small
\centering
\begin{tabular}{lccccccccccc}
\toprule
\multirow{2}{*}{\textbf{dataset}} & \multirow{2}{*}{\textbf{prompt}} & \multicolumn{2}{c}{GPT-JT-6B} & \multicolumn{2}{c}{LLaMA2-13B-chat} & \multicolumn{2}{c}{LLaMA2-7B-chat} & \multicolumn{2}{c}{Flan-T5-11B} & \multicolumn{2}{c}{Flan-T5-770M} \\ \cmidrule{3-12}
                      &                  & \textit{sad}                & \textit{greedy} & \textit{sad}    & \textit{greedy}  & \textit{sad}   & \textit{greedy} & \textit{sad}     & \textit{greedy}  & \textit{sad}     & \textit{greedy}   \\ \midrule
\multirow{3}{*}{CoLA} & \texttt{base\_a} & \cellcolor{green}45.16 & 38.71       & 59.20                       & 59.20      & 52.37                      & 52.56 & 29.41                       & 29.41 & \cellcolor{green}41.75 & 40.99 \\
                      & \texttt{base\_b} & \cellcolor{green}48.01 & 47.82       & 70.21                       & 70.40 & 62.62                       & 63.38 & 77.42                       & 81.59 & 71.54                       & 71.92 \\
                      & \texttt{CFP}     & \cellcolor{green}38.71 & 34.91       & \cellcolor{green}67.93 & 67.74      & 68.31                       & 68.88 & 77.61                       & 80.65 & 69.45                       & 69.45      \\ \midrule
\multirow{3}{*}{CSQA} & \texttt{base\_a} & 26.48                       & 26.56  & 45.49                       & 46.80 & 33.20                       & 33.93 & \cellcolor{green}57.79 & 55.82      & 79.02                       & 83.77 \\
                      & \texttt{base\_b} & 30.74                       & 30.74       & 61.97                       & 62.13 & 54.59                       & 55.49 & 78.85                       & 82.30 & 24.92                       & 25.41 \\
                      & \texttt{CFP}     & 27.70                       & 28.20  & 60.98                       & 61.72 & \cellcolor{green}54.59 & 54.51      & 77.30                       & 81.64 & 22.79                       & 22.87 \\ \midrule
\multirow{3}{*}{MNLI} & \texttt{base\_a} & \cellcolor{green}14.90 & 11.50       & \cellcolor{green}9.20  & 3.40       & 16.90                       & 17.20 & 23.50                       & 25.80 & \cellcolor{green}9.10  & 3.70       \\
                      & \texttt{base\_b} & \cellcolor{green}27.30 & 27.00       & \cellcolor{green}34.30 & 30.30      & \cellcolor{green}37.00 & 34.10      & 81.60                       & 82.30 & 59.80                       & 65.10 \\
                      & \texttt{CFP}     & 33.50                       & 34.20  & \cellcolor{green}43.90 & 39.70      & 39.90                       & 40.40 & 81.40                       & 82.90 & \cellcolor{green}14.60 & 9.10       \\ \midrule
\multirow{3}{*}{RTE}  & \texttt{base\_a} & \cellcolor{green}37.55 & 31.05       & 38.99                       & 39.71 & \cellcolor{green}34.66 & 32.85      & \cellcolor{green}24.19 & 23.83      & \cellcolor{green}22.38 & 21.66      \\
                      & \texttt{base\_b} & 58.12                       & 58.48  & \cellcolor{green}53.07 & 37.55      & 62.82                       & 63.18 & 82.31                       & 87.00 & 50.54                       & 54.15 \\
                      & \texttt{CFP}     & \cellcolor{green}52.71 & 50.18       & \cellcolor{green}56.32 & 43.68      & 56.32                       & 56.32      & 75.09                       & 85.56 & 44.77                       & 53.79 \\ \midrule
\multirow{3}{*}{SST2} & \texttt{base\_a} & 25.34                       & 27.29  & \cellcolor{green}59.98 & 59.52      & 61.12                       & 61.47 & 50.80                       & 50.80      & 40.14                       & 40.94 \\
                      & \texttt{base\_b} & 76.95                       & 78.44  & \cellcolor{green}77.64 & 77.41      & 78.10                       & 78.44 & 90.02                       & 95.64 & 91.63                       & 91.97 \\
                      & \texttt{CFP}     & 53.56                       & 54.13  & 88.07                       & 88.19 & \cellcolor{green}77.06 & 76.49      & 91.17                       & 95.87 & 77.87                       &78.10 \\
\bottomrule
\end{tabular}
\caption{The highest accuracy reached using \textit{sensitivity-aware} decoding (\textit{sad}) and the accuracy of greedy decoding (\textit{greedy}). Cases where \textit{sensitivity-aware} decoding has a better accuracy than greedy decoding are \sethlcolor{green}\hl{highlighted}.}
\label{best_sad_and_gd_comprison}
\end{table*}

\textit{Prompt} tokens are important in the sense that they provide information such as instructions and external knowledge, which are necessary for the model to produce outputs that the user expects. However, it is counter-intuitive that the \textit{mean saliency scores} for \textit{prompt} tokens are much higher than those for \textit{input} tokens, as \textit{input} tokens contain essential information as well. This observation may imply that memory plays an overwhelming role in ICL \cite{chen-etal-2022-improving,merullo2023mechanism,mckenna2023sources,tefnik-kadlcik-2023-context,singh2023transient}--LLMs were trained on similar instances, so they do not need to rely on \textit{input} tokens in the test instances too much. In this sense, \textit{prompt} tokens are more relevant because they trigger memories, and models implicitly infer task information from them \cite{reynolds2021prompt,hendel2023incontext,wang2023llms,wolf2023fundamental}. Over-emphasis on \textit{prompt} tokens may lead to hallucination as well. A recent study discovers that instruction tuning significantly increases sycophancy in LLMs, that they follow user’s opinion or agree with user's claim even when they know it is false \cite{wei2023simple}.

\section{\textit{Sensitivity-aware} decoding}

We showed in Section \ref{results} that sensitivity can be viewed as an unsupervised proxy of accuracy. In this section, we further show that including sensitivity in decoding improves model performance. Specifically, we add sensitivity as a penalty to greedy decoding:
\begin{equation}
    \hat{y} = \argmax_{y \in V}[\alpha P(y|x) - (1-\alpha) s],
\end{equation}
\noindent where $x$ is an input, $V$ is a vocabulary, and $\hat{y}$ is the output. $P(y|x)$ is the probability of an output $y$ given $x$, and $s$ is a sensitivity estimation, calculated as the variance of the output logits of the synthetic data for $x$. We reweight $P(y|x)$ and $s$ using $\alpha$ and $(1-\alpha)$.

Table \ref{best_sad_and_gd_comprison} summarizes the performance of \textit{sensitivity-aware} decoding compared to greedy decoding. \textit{Sensitivity-aware} decoding works better on CoLA, MNLI, and RTE, and model-wisely, it works better with GPT-JT-6B and LLaMA2-13B-chat. \textit{Sensitivity-aware} decoding works much better with \texttt{base\_a} than the other prompts. See \ref{more_on_sensitivity_aware_decoding} for more implementation details and results.

The results show that penalizing outputs with high sensitivity in decoding has an effect on model performance. We show that \textit{sensitivity-aware} decoding works better with the most basic prompt, \texttt{base\_a}, that contains plain \textit{input-target} pairs. We believe this prompt represents those truly challenging problems in real life that have almost no clues or hints, and require very strong reasoning abilities to solve. Apparently, none of the models we tested possess such abilities. Therefore, \textit{sensitivity-aware} decoding, which helps improve model performance under this ``extreme condition,'' is highly meaningful in the present context.

However, \textit{sensitivity-aware} decoding is more computationally expensive compared to standard greedy decoding as it requires multiple inference passes. This makes it impractical for tasks that demand low latency.





\section{Conclusion}

This work provides a novel perspective on prompt analysis, examining the effects of prompts in terms of the sensitivity of a function. We conducted a systematic and comprehensive analysis, and highlight how certain prompts are more effective due to their ability to reduce sensitivity levels. We show that sensitivity can serve as an unsupervised proxy of model performance, making it a valuable tool for evaluating model performance without using labeled data or ground truth. By introducing \textit{sensitivity-aware} decoding, we show that incorporating sensitivity in greedy decoding is particularly helpful in cases where the input is less informative. Since none of the models we tested performs well when there is limited information in the input, we believe \textit{sensitivity-aware} decoding is highly practical in the current context. Our work not only sheds light on prompt engineering, but also provides insight into the working mechanism of ICL.

\clearpage

\section*{Limitations}

The measurement of sensitivity is currently quite restricted to close-ended generation. It is challenging to extend this framework to tasks such as text summarization. While we demonstrate the effectiveness of \textit{sensitivity-aware} decoding, it requires multiple inferences, which may be impractical for tasks that require low latency. In order to manage costs, we limited the use of OpenAI \texttt{text-davinci-003} (GPT3.5-175B) in our experiments. Due to its closed source nature, the reproducibility of the results related to GPT3.5-175B may be a concern as well.

\section*{Acknowledgements}

This research work has been funded by the German Federal Ministry of Education and Research and the Hessian Ministry of Higher Education, Research, Science and the Arts within their joint support of the National Research Center for Applied Cybersecurity ATHENE.

\bibliography{anthology,custom}
\bibliographystyle{acl_natbib}

\clearpage

\onecolumn
\appendix
\section{Appendix}
\label{appendix}

\subsection{More on experiment settings}
\label{more_on_experiment_settings}

All experiments were done in the few-shot setting.\footnote{See \url{https://github.com/UKPLab/naacl2024-prompt-sensitivity/tree/main/prompts}.} We set \texttt{temperature=0.8} for all experiments. We set \texttt{max\_new\_tokens=2}, except for experiments with \texttt{CoT}, where it was set to 64, and 128 for experiments with GSM8K. For local models, we set \texttt{batch\_size=16} and \texttt{seeds=[2266,105,86379]}, and all experiments were run on either an NVIDIA A100 or H100. The experiments with OpenAI \texttt{text-davinci-003} were done between July 8 and July 18, 2023.

We notice that the synthetic data for CSQA are noisier than those for other datasets. We manually checked the synthetic data for the first 50 instances in CSQA (a total of 1220 instances). There are 223 synthetic data for the first 50 instances, among which 44 are considered to be noisy (see Table \ref{noisy_data_csqa}).

We designed \texttt{base\_a}, \texttt{base\_b}, \texttt{zero\_a}, and \texttt{zero\_b} in an intuitive way. We did not rely on any formal theories or guidelines related to prompt engineering. Table \ref{demonstrations} shows examples of demonstrations with different prompts.

\begin{small}
\begin{longtable}{p{0.46\linewidth}p{0.46\linewidth}}
\endfirsthead
\toprule
\textbf{original} & \textbf{synthetic} \\ \midrule
\endhead
\toprule
\textbf{original} & \textbf{synthetic} \\ \midrule
Where would you find magazines along side many other printed works?                                                   & \multirow{2}{*}{:-Where would you waste your Spanish ?? !?}                                                                    \\ \midrule
Where are  you likely to find a hamburger?                                                                            & prime over byEL are you "happy" hamburger?                                                                                     \\ \midrule
\multirow{2}{\linewidth}{James was looking for a good place to buy farmland. Where might he look?}                    & ei-- What look is for a good place to buy -- like a restaurant?                                                                \\ \cmidrule{2-2}
                                                                                                                      & <..."What might he look?                                                                                                       \\ \midrule
\multirow{4}{\linewidth}{In what Spanish speaking North American country can you get a great cup of coffee?}          & Balkan Kingdom \textless Meditation a French speaking North American country can you explain the ways components are used?     \\ \cmidrule{2-2}
                                                                                                                      & ///5// Does insurance cover a car, Sears, or a cup of coffee?                                                                  \\ \midrule
\multirow{4}{\linewidth}{What do animals do when an enemy is approaching?}                                            & 19:: What do animals do when an enemy is spotted?                                                                              \\ \cmidrule{2-2}
                                                                                                                      & com... What is an enemy of HTML?                                                                                               \\ \cmidrule{2-2}
                                                                                                                      & p-0 What do you do when an enemy moves in?                                                                                     \\ \midrule
\multirow{2}{\linewidth}{Reading newspaper one of many ways to practice your what?}                                   & =- Your WHERE? Your ARE. Your what?                                                                                            \\ \cmidrule{2-2}
                                                                                                                      & an tissues/Reading newspaper/What is a red/aloe?                                                                               \\ \midrule
If you want harmony, what is something you should try to do with the world?                                           & Sept: 5If you want to have a   process then what do you now do with the system?                                                \\ \midrule
\multirow{2}{\linewidth}{Where does a heifer's master live?}                                                          & =- does a heifer take in valuables?                                                                                            \\ \cmidrule{2-2}
                                                                                                                      & /Jazeeraio "What's a Maru?                                                                                                     \\ \midrule
Aside from water and nourishment what does your dog need?                                                             & \multirow{2}{*}{/ Purity Utah Prison of 2015?}                                                                                 \\ \midrule
\multirow{2}{\linewidth}{Janet was watching the film because she liked what?}                                         & ///Janet was watching the film because she liked what?                                                                         \\ \cmidrule{2-2}
                                                                                                                      & 08/Janet was watching the film "Sha what?                                                                                      \\ \midrule
\multirow{4}{\linewidth}{What are you waiting alongside with when you're in a reception area?}                        & 27.9What are you waiting alongside with when you're in a job interview?                                                        \\ \cmidrule{2-2}
                                                                                                                      & 1: $\boxtimes \boxtimes$ hat are you going to do while on the street a day like now?                                           \\ \midrule
\multirow{4}{\linewidth}{When drinking booze what can you do to stay busy?}                                           & When drinking : When more can you do then pleasure ?                                                                           \\ \cmidrule{2-2}
                                                                                                                      & When drinking ? When being a kid/ Young man ?                                                                                  \\ \cmidrule{2-2}
                                                                                                                      & =-Bin Bak Allah? Bern Abd Ja Tih?                                                                                              \\ \midrule
\multirow{4}{\linewidth}{A fencing thrust with a sharp sword towards a person would result in what?}                  & 2, 45A fencing thrust with a sharp sword towards a person would result in pain?                                                \\ \cmidrule{2-2}
                                                                                                                      & Video Audio AudioA fencing thrust from the guitarist/band member whether way or the other?                                     \\ \midrule
Unlike a spider and his many sight seers, people only have what?                                                      & ") cards Of the spider spider and his many sight seers. We Who See Everything?                                                 \\ \midrule
What could go on top of wood?                                                                                         & \{50 What you could go wrong with this title?                                                                                  \\ \midrule
The artist was sitting quietly pondering, then suddenly he began to paint when   what struck him?                     & =- Imagine an artist was sitting quietly pondering, then taking a break to paint when what struck him?                         \\ \midrule
\multirow{2}{\linewidth}{Where could you find a toilet that only friends can use?}                                    & =---Where could you find a toilet maker with a slot underneath?                                                                \\ \midrule
\multirow{2}{\linewidth}{What is someone who isn't clever, bright, or competent called?}                              & 1. Why are aren $\boxtimes \boxtimes$ clever, bright, or competent called?                                                     \\ \cmidrule{2-2}
                                                                                                                      & s// ? Him who isn't clever, ?   peculiar, ?                                                                                    \\ \midrule
\multirow{5}{\linewidth}{When wildlife reproduce we often refer to what comes out as what?}                           & ( Do we often refer to wha  comes out as what?                                                                                 \\ \cmidrule{2-2}
                                                                                                                      & div 2: were you aware we often refer to what comes out as what?                                                                \\ \cmidrule{2-2}
                                                                                                                      & ",.."?                                                                                                                         \\ \midrule
Blue read material outside of his comfort zone because he wanted to gain what?                                        & //Blue read material/ text/   flag/ license/ licensing/ gain what?                                                             \\ \midrule
After he got hired he hoped for success at his what?                                                                  & /2011-2012-2013. Who is his what?                                                                                              \\ \midrule
\multirow{2}{\linewidth}{Committing perjury is a serious what?}                                                       & ned!? $\boxtimes$?perjury is involved here??                                                                                   \\ \cmidrule{2-2}
                                                                                                                      & ///committing test ? Now what?                                                                                                 \\ \midrule
The lock kept the steering wheel from moving, but the thief still took his chances and   began to work on the what?   & =-- Maybe one lock kept the steering wheel from moving, but the thief still had the gun. If not, what was it?                  \\ \midrule
Who is a police officer likely to work for?                                                                           & based? Do you consider them likely to be sustainable?                                                                          \\ \midrule
\multirow{2}{\linewidth}{Where is a doormat likely to be in front of?}                                                & Pop RSWhere is a substitute to an already terminated system in comparison?                                                     \\ \midrule
He needed more information to fix it, so he consulted the what?                                                       & 1. 13He needed more information to fix information, but defacto what?                                                          \\ \midrule
\multirow{4}{\linewidth}{Where can you put a picture frame when it's not hung vertically?}                            & \%\%\%Where can you put a picture frame when it's not nesting?                                                                 \\ \cmidrule{2-2}
                                                                                                                      & min "Where can you be if your wife or any other object is hung vertically?                                                     \\ \midrule
What must someone do before they shop?	                                                                              & /// How must someone conduct their work daily?                                                                                 \\
\bottomrule
\caption{Noisy synthetic data for the first 50 instances in CSQA.}
\label{noisy_data_csqa}
\end{longtable}
\end{small}

\clearpage

\begin{small}
\begin{longtable}{lcp{0.72\linewidth}}
\endfirsthead
\toprule
\textbf{dataset} & \textbf{prompt} & \textbf{text} \\ \midrule
\endhead
\toprule
\textbf{dataset} & \textbf{prompt} & \textbf{text}                                                                                                    \\ \midrule
\multirow{27}{*}{CSQA} & \multirow{3}{*}{\texttt{base\_a}} & Google Maps and other highway and street GPS services have replaced what?                               \\
                       &                                   & (0) united states, (1) mexico, (2) countryside, (3) atlas, (4) oceans                                   \\
                       &                                   & 3                                                                                                       \\ \cmidrule{2-3}
                       & \multirow{3}{*}{\texttt{base\_b}} & \textsc{sentence:} Google Maps and other highway and street GPS services have replaced what?                     \\
                       &                                   & \textsc{question:} Is it (0) united states, (1) mexico, (2) countryside, (3) atlas, (4) oceans?                  \\
                       &                                   & \textsc{answer:} 3                                                                                               \\ \cmidrule{2-3}
                       & \multirow{4}{*}{\texttt{zero\_a}} & Google Maps and other highway and street GPS services have replaced what?                               \\
                       &                                   & (0) united states, (1) mexico, (2) countryside, (3) atlas, (4) oceans                                   \\
                       &                                   & The answer is 3.                                                                                        \\
                       &                                   & 3                                                                                                       \\ \cmidrule{2-3}
                       & \multirow{4}{*}{\texttt{zero\_b}} & \textsc{sentence:} Google Maps and other highway and street GPS services have replaced what?                     \\
                       &                                   & \textsc{options:} (0) united states, (1) mexico, (2) countryside, (3) atlas, (4) oceans                          \\
                       &                                   & The answer is 3.                                                                                        \\
                       &                                   & \textsc{answer:} 3                                                                                               \\ \cmidrule{2-3}
                       & \multirow{4}{*}{\texttt{CFP}}     & Bob said, "Google Maps and other highway and street GPS services have replaced what?"                   \\
                       &                                   & \textsc{question:} Is it (0) united states, (1) mexico, (2) countryside, (3) atlas, (4) oceans in Bob's opinion? \\
                       &                                   & \textsc{answer:} 3                                                                                               \\ \cmidrule{2-3}
                       & \multirow{5}{*}{\texttt{CoT}}     & \textsc{sentence:} Google Maps and other highway and street GPS services have replaced what?                     \\
                       &                                   & \textsc{question:} Is it (0) united states, (1) mexico, (2) countryside, (3) atlas, (4) oceans?                  \\
                       &                                   & \textsc{answer:} Let's think step by step. Google Maps and other highway and street GPS services help people find their location and navigate streets and highways, and atlas is a software that is designed to help users navigate. So the answer is 3.  \\ \cmidrule{2-3}
                       & \multirow{4}{*}{\texttt{GKP}}     & \textsc{knowledge:} Electronic maps are the modern version of paper atlas. \\
                       &                                   & \textsc{input:} Google Maps and other highway and street GPS services have replaced what? \\
                       &                                   & \textsc{options:} (0) united states, (1) mexico, (2) countryside, (3) atlas, (4) oceans \\
                       &                                   & \textsc{output:} 3 \\ \midrule
\multirow{38}{*}{MNLI} & \multirow{5}{*}{\texttt{base\_a}} & He thought about ways to achieve this life goal for a long time, which means until he learned the basics of text editing, which happened at his first job at a firm trading in plastic bags landfill disposal permits.                                                                                                                                \\
                       &                          & He thought about ways to achieve his life goals for a long time.                                                                                                                                                                                                                                                                                      \\
                       &                          & 1                                                                                                                                                                                                                                                                                                                                                     \\ \cmidrule{2-3}
                       & \multirow{5}{*}{\texttt{base\_b}} & \textsc{sentence1:} He thought about ways to achieve this life goal for a long time, which means until he learned the basics of text editing, which happened at his first job at a firm trading in plastic bags landfill disposal permits.                                                                                                                     \\
                       &                          & \textsc{sentence2:} He thought about ways to achieve his life goals for a long time.                                                                                                                                                                                                                                                                           \\
                       &                          & \textsc{answer:} 1                                                                                                                                                                                                                                                                                                                                             \\ \cmidrule{2-3}
                       & \multirow{6}{*}{\texttt{zero\_a}} & He thought about ways to achieve this life goal for a long time, which means until he learned the basics of text editing, which happened at his first job at a firm trading in plastic bags landfill disposal permits.                                                                                                                                \\
                       &                          & He thought about ways to achieve his life goals for a long time.                                                                                                                                                                                                                                                                                      \\
                       &                          & The answer is 1.                                                                                                                                                                                                                                                                                                                                      \\
                       &                          & 1                                                                                                                                                                                                                                                                                                                                                     \\ \cmidrule{2-3}
                       & \multirow{6}{*}{\texttt{zero\_b}} & \textsc{sentence1:} He thought about ways to achieve this life goal for a long time, which means until he learned the basics of text editing, which happened at his first job at a firm trading in plastic bags landfill disposal permits.                                                                                                                     \\
                       &                          & \textsc{sentence2:} He thought about ways to achieve his life goals for a long time.                                                                                                                                                                                                                                                                           \\
                       &                          & The answer is 1.                                                                                                                                                                                                                                                                                                                                      \\
                       &                          & \textsc{answer:} 1                                                                                                                                                                                                                                                                                                                                             \\ \cmidrule{2-3}
                       & \multirow{7}{*}{\texttt{CFP}}     & Bob said, "sentence 1 is 'He thought about ways to achieve this life goal for a long time, which means until he learned the basics of text editing, which happened at his first job at a firm trading in plastic bags landfill disposal permits,' and sentence 2 is 'He thought about ways to achieve his life goals for a long time.'"               \\
                       &                          & \textsc{question:} Are the two sentences (0) contradiction, (1) entailment, or (2) neutral in Bob's opinion?                                                                                                                                                                                                                                                   \\
                       &                          & \textsc{answer:} 1                                                                                                                                                                                                                                                                                                                                             \\ \cmidrule{2-3}
                       & \multirow{9}{*}{\texttt{CoT}}     & \textsc{sentence1:} He thought about ways to achieve this life goal for a long time, which means until he learned the basics of text editing, which happened at his first job at a firm trading in plastic bags landfill disposal permits.                                                                                                                     \\
                       &                          & \textsc{sentence2:} He thought about ways to achieve his life goals for a long time.                                                                                                                                                                                                                                                                           \\
                       &                          & \textsc{question:} Is this (0) contradiction, (1) entailment, or (2) neutral?                                                                                                                                                                                                                                                                                  \\
                       &                          & \textsc{answer:} Let's think step by step. Sentence 1 states that he thought about ways to achieve his life goal for a long time, and then states that he learned the basics of text editing at his first job. Sentence 2 states that he thought about ways to achieve his life goal for a long time. Both sentences state the same thing, so the answer is 1.  \\ \midrule
\multirow{49}{*}{RTE} & \multirow{5}{*}{\texttt{base\_a}} & The Federal Bureau of Investigation started an independent probe of the circumstances shortly after the White House made plain that President Bill Clinton considered industrial espionage a particular threat to US economic interests.                                                                                                                                          \\
                      &                          & President Clinton thinks that industrial espionage is a threat to America's well being.                                                                                                                                                                                                                                                                                           \\
                      &                          & 0                                                                                                                                                                                                                                                                                                                                                                                 \\ \cmidrule{2-3}
                      & \multirow{7}{*}{\texttt{base\_b}} & \textsc{sentence1:} The Federal Bureau of Investigation started an independent probe of the circumstances shortly after the White House made plain that President Bill Clinton considered industrial espionage a particular threat to US economic interests.                                                                                                                               \\
                      &                          & \textsc{sentence2:} President Clinton thinks that industrial espionage is a threat to America's well being.                                                                                                                                                                                                                                                                                \\
                      &                          & \textsc{question:} Are the two sentences (0) entailment, or (1) not\_entailment?                                                                                                                                                                                                                                                                                                           \\
                      &                          & \textsc{answer:} 0                                                                                                                                                                                                                                                                                                                                                                         \\ \cmidrule{2-3}
                      & \multirow{6}{*}{\texttt{zero\_a}} & The Federal Bureau of Investigation started an independent probe of the circumstances shortly after the White House made plain that President Bill Clinton considered industrial espionage a particular threat to US economic interests.                                                                                                                                          \\
                      &                          & President Clinton thinks that industrial espionage is a threat to America's well being.                                                                                                                                                                                                                                                                                           \\
                      &                          & The answer is 0.                                                                                                                                                                                                                                                                                                                                                                  \\
                      &                          & 0                                                                                                                                                                                                                                                                                                                                                                                 \\ \cmidrule{2-3}
                      & \multirow{7}{*}{\texttt{zero\_b}} & \textsc{sentence1:} The Federal Bureau of Investigation started an independent probe of the circumstances shortly after the White House made plain that President Bill Clinton considered industrial espionage a particular threat to US economic interests.                                                                                                                               \\
                      &                          & \textsc{sentence2:} President Clinton thinks that industrial espionage is a threat to America's well being.                                                                                                                                                                                                                                                                                \\
                      &                          & The answer is 0.                                                                                                                                                                                                                                                                                                                                                                  \\
                      &                          & \textsc{answer:} 0                                                                                                                                                                                                                                                                                                                                                                         \\ \cmidrule{2-3}
                      & \multirow{6}{*}{\texttt{CFP}}     & Bob said, "sentence 1 is 'The Federal Bureau of Investigation started an independent probe of the circumstances shortly after the White House made plain that President Bill Clinton considered industrial espionage a particular threat to US economic interests.', and sentence 2 is 'President Clinton thinks that industrial espionage is a threat to America's well being.'" \\
                      &                          & \textsc{question:} Are the two sentences (0) entailment, or (1) not\_entailment in Bob's opinion?                                                                                                                                                                                                                                                                                          \\
                      &                          & \textsc{answer:} 0                                                                                                                                                                                                                                                                                                                                                                         \\ \cmidrule{2-3}
                      & \multirow{10}{*}{\texttt{CoT}}     & \textsc{sentence1:} The Federal Bureau of Investigation started an independent probe of the circumstances shortly after the White House made plain that President Bill Clinton considered industrial espionage a particular threat to US economic interests.                                                                                                                               \\
                      &                          & \textsc{sentence2:} President Clinton thinks that industrial espionage is a threat to America's well being.                                                                                                                                                                                                                                                                                \\
                      &                          & \textsc{question:} Are the two sentences (0) entailment, or (1) not\_entailment?                                                                                                                                                                                                                                                                                                           \\
                      &                          & \textsc{answer:} Let's think step by step. The first sentence provides information about President Clinton's belief, which directly leads to the FBI conducting an investigation. Therefore, the second sentence necessarily follows the information provided in the first sentence. So the answer is 0.                                                                                   \\ \cmidrule{2-3}
                      & \multirow{8}{*}{\texttt{APE}}     & \textsc{instruction:} determine whether the given statement logically follows from the preceding statement, and the output is either (0) entailment if the statement logically follows, or (1) not\_entailment if the statement does not logically follow.                                                                                                                                 \\
                      &                          & \textsc{input:} The Federal Bureau of Investigation started an independent probe of the circumstances shortly after the White House made plain that President Bill Clinton considered industrial espionage a particular threat to US economic interests.                                                                                                                                   \\
                      &                          & President Clinton thinks that industrial espionage is a threat to America's well being.                                                                                                                                                                                                                                                                                           \\
                      &                          & \textsc{output:} 0  \\ \midrule
\multirow{25}{*}{SST2} & \multirow{3}{*}{\texttt{base\_a}} & None of this is meaningful or memorable , but frosting is n't , either , and you would n't turn down a big bowl of that , would you ?                                                                          \\
                       &                          & 0                                                                                                                                                                                                              \\ \cmidrule{2-3}
                       & \multirow{4}{*}{\texttt{base\_b}} & \textsc{sentence:} None of this is meaningful or memorable , but frosting is n't , either , and you would n't turn down a big bowl of that , would you ?                                                                \\
                       &                          & \textsc{question:} Is this (0) negative, or (1) positive?                                                                                                                                                               \\
                       &                          & \textsc{answer:} 0                                                                                                                                                                                                      \\ \cmidrule{2-3}
                       & \multirow{4}{*}{\texttt{zero\_a}} & None of this is meaningful or memorable , but frosting is n't , either , and you would n't turn down a big bowl of that , would you ?                                                                          \\
                       &                          & The answer is 0.                                                                                                                                                                                               \\
                       &                          & 0                                                                                                                                                                                                              \\ \cmidrule{2-3}
                       & \multirow{4}{*}{\texttt{zero\_b}} & \textsc{sentence:} None of this is meaningful or memorable , but frosting is n't , either , and you would n't turn down a big bowl of that , would you ?                                                                \\
                       &                          & The answer is 0.                                                                                                                                                                                               \\
                       &                          & \textsc{answer:} 0                                                                                                                                                                                                      \\ \cmidrule{2-3}
                       & \multirow{4}{*}{\texttt{CFP}}     & Bob said, "None of this is meaningful or memorable , but frosting is n't , either , and you would n't turn down a big bowl of that , would you ?"                                                              \\
                       &                          & \textsc{question:} Is this (0) negative, or (1) positive in Bob's opinion?                                                                                                                                              \\
                       &                          & \textsc{answer:} 0                                                                                                                                                                                                      \\ \cmidrule{2-3}
                       & \multirow{6}{*}{\texttt{CoT}}     & \textsc{sentence:} None of this is meaningful or memorable , but frosting is n't , either , and you would n't turn down a big bowl of that , would you ?                                                                \\
                       &                          & \textsc{question:} Is this (0) negative, or (1) positive?                                                                                                                                                               \\
                       &                          & \textsc{answer:} Let's think step by step. "None of this is meaningful or memorable" suggests that whatever is being referred to is not significant or noteworthy, indicating a negative sentiment. So the answer is 0. \\ \midrule
\multirow{19}{*}{GSM8K} & \multirow{5}{*}{\texttt{base\_b}} & \textsc{sentence:} Sally and Bob have made plans to go on a trip at the end of the year. They both decide to work as babysitters and save half of what they've earned for their trip. If Sally makes \$6 per day and Bob makes \$4 per day, how much money will they both have saved for their trip after a year? \\
                       &                          & \textsc{answer:} 1825 \\ \cmidrule{2-3}
                       & \multirow{5}{*}{\texttt{zero\_b}} & \textsc{sentence:} Sally and Bob have made plans to go on a trip at the end of the year. They both decide to work as babysitters and save half of what they've earned for their trip. If Sally makes \$6 per day and Bob makes \$4 per day, how much money will they both have saved for their trip after a year? The answer is 1825. \\
                       &                          & \textsc{answer:} 1825 \\ \cmidrule{2-3}
                       & \multirow{9}{*}{\texttt{CoT}} & \textsc{sentence:} Sally and Bob have made plans to go on a trip at the end of the year. They both decide to work as babysitters and save half of what they've earned for their trip. If Sally makes \$6 per day and Bob makes \$4 per day, how much money will they both have saved for their trip after a year? \\
                       &                          & \textsc{answer:} Let's think step by step. Saly saves 1/2 * \$6/day = 3/day. Since each year have 365 days, the total amount of money Sally will save in a year is \$3/day * 365 days/year = 1095/year. Bob saves 1/2 * \$4/day = 2/day. The total amount of money Bob will have saved in a year is \$2/day * 365 days/year = 730/year In total, Sally and Bob would have saved \$730 + \$1095 = 1825. So the answer is 1825. \\
\bottomrule
\caption{Examples of demonstrations with different prompts. Demonstrations for CoLA are exemplified in Table \ref{prompt_reference}, so they are not included here.}
\label{demonstrations}
\end{longtable}
\end{small}

\clearpage

\subsection{More on LLaMA results}
LLaMA2 models were not available when we began our experiments. As a result, we conducted our initial experiments using LLaMA-7B, LLaMA-13B, and LLaMA-30B \cite{touvron2023llama}. Table \ref{llama_accuracy_sensitivity} shows a comparison between LLaMA2 and LLaMA models.

\begin{table}[H]
\small
\centering
\begin{tabular}{lcc}
\toprule
\textbf{model} & \textbf{accuracy$\uparrow$} & \textbf{sensitivity$\downarrow$} \\ \midrule
LLaMA2-13B-chat   & 0.6961 & 0.1883 \\
LLaMA2-7B-chat    & 0.6843 & 0.1951 \\
LLaMA-30B         & 0.6835 & 0.2324 \\
LLaMA-13B         & 0.6177 & 0.2327 \\
LLaMA-7B          & 0.6010 & 0.2457 \\
\bottomrule
\end{tabular}
\caption{The average accuracy and sensitivity of LLaMA models across different tasks.}
\label{llama_accuracy_sensitivity}
\end{table}


\subsection{More on instruction, \textit{chain-of-thought}, and knowledge}
\label{more_on_instruction_chain_of_thought_and_knowledge}

Table \ref{base_a_and_base_b} shows the average performance across different models using \texttt{base\_a} and \texttt{base\_b}. Table \ref{prompt_reference_cot_standard_a} shows examples of \texttt{CoT\_base\_a}.

\begin{table}[H]
\small
\centering
\begin{tabular}{lccccc}
\toprule
\textbf{dataset}      & \textbf{prompt}          & \textbf{accuracy} & \textbf{sensitivity} & \textbf{accuracy +/-}       & \textbf{sensitivity +/-}      \\ \midrule
\multirow{2}{*}{CoLA} & \texttt{base\_a}         & 0.4215        & 0.3317         & \multirow{2}{*}{0.2275} & \multirow{2}{*}{0.1125} \\
                      & \texttt{base\_b}         & 0.6490        & 0.2192         &                         &                         \\ \midrule
\multirow{2}{*}{CSQA} & \texttt{base\_a}         & 0.4652        & 0.4107         & \multirow{2}{*}{0.0235} & \multirow{2}{*}{0.0587} \\
                      & \texttt{base\_b}         & 0.4887        & 0.3520         &                         &                         \\ \midrule
\multirow{2}{*}{MNLI} & \texttt{base\_a}         & 0.1317        & 0.4498         & \multirow{2}{*}{0.3258} & \multirow{2}{*}{0.1405} \\
                      & \texttt{base\_b}         & 0.4575        & 0.3094         &                         &                         \\ \midrule
\multirow{2}{*}{RTE}  & \texttt{base\_a}         & 0.2332        & 0.5120         & \multirow{2}{*}{0.3487} & \multirow{2}{*}{0.2763} \\
                      & \texttt{base\_b}         & 0.5819        & 0.2357         &                         &                         \\ \midrule
\multirow{2}{*}{SST2} & \texttt{base\_a}         & 0.3841        & 0.4198         & \multirow{2}{*}{0.3992} & \multirow{2}{*}{0.2318} \\
                      & \texttt{base\_b}         & 0.7833        & 0.1880         &                         &                         \\
\bottomrule
\end{tabular}
\caption{The average accuracy and sensitivity across different models, and the difference between the accuracy (\textbf{accuracy +/-}) and sensitivity (\textbf{sensitivity +/-}) of models using \texttt{base\_a} and \texttt{base\_b}.}
\label{base_a_and_base_b}
\end{table}

\begin{table}[H]
\small
\centering
\begin{tabular}{lp{0.68\linewidth}}
\toprule
\textbf{dataset} & \textbf{text} \\ \midrule
\multirow{3}{*}{CoLA} & I'm glad I saw anybody. \\
& Let's think step by step. This sentence is ungrammatical because ``anybody'' is used as the object in an affirmative clause. So the answer is 0. \\ \midrule
\multirow{8}{*}{RTE} & The Federal Bureau of Investigation started an independent probe of the circumstances shortly after the White House made plain that President Bill Clinton considered industrial espionage a particular threat to US economic interests. \\
& President Clinton thinks that industrial espionage is a threat to America's well being. \\
& Let's think step by step. The first sentence provides information about President Clinton's belief, which directly leads to the FBI conducting an investigation. Therefore, the second sentence follows the information provided in the first sentence. So the answer is 0. \\
\bottomrule
\end{tabular}
\caption{Examples of \texttt{CoT\_base\_a} for CoLA and RTE.}
\label{prompt_reference_cot_standard_a}
\end{table}

\subsection{More on Flan-T5 with \texttt{zero}}
\label{more_on_flan_t5_with_zero}

\begin{figure}[H]
\centering
\includegraphics[width=6.8cm]{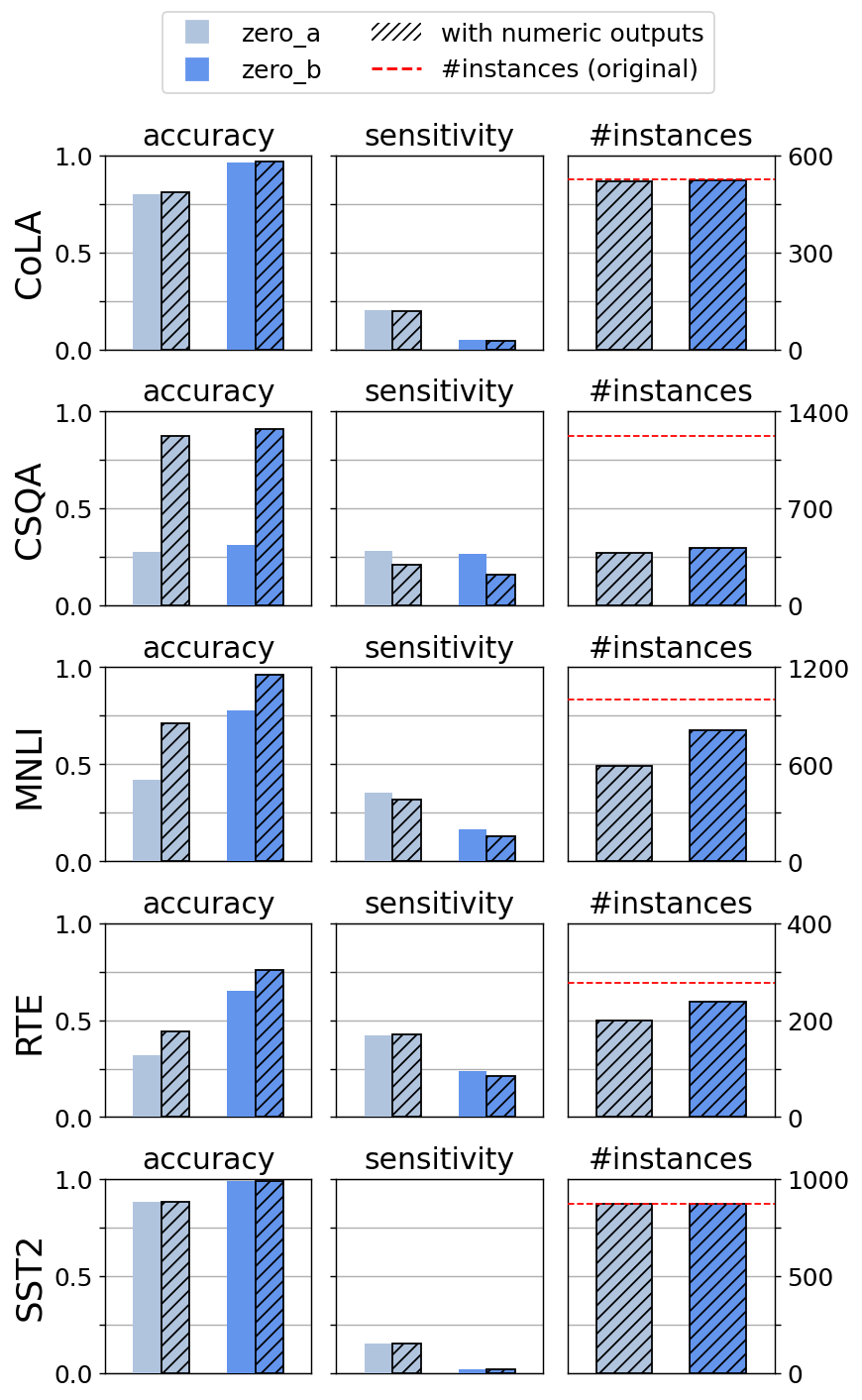}
\caption{The accuracy and sensitivity of Flan-T5-11B on the full dataset using \texttt{zero\_a} and \texttt{zero\_b}, and the accuracy and sensitivity of Flan-T5-11B on instances where it outputs a numeric index as what is exemplified in the prompt. \texttt{\#instances} shows the number of instances in a dataset (\texttt{\#instances (original)}) and the number of instances where Flan-T5-11B outputs a numeric index (\texttt{with numeric outputs}).}
\label{flan_t5_with_zero}
\end{figure}

Figure \ref{flan_t5_with_zero} shows the performance of Flan-T5-11B using the \texttt{zero} prompts. Flan-T5-11B fails to output a numeric index on CSQA, MNLI, and RTE instances in most cases. Table \ref{t5_zero_b} shows the performance of Flan-T5 models using \texttt{zero\_b}. Table \ref{t5_with_base_a} shows the original number of instances in each dataset and the number of instances where Flan-T5 models output a numeric index using \texttt{base\_a}.

\begin{table}[H]
\small
\centering
\begin{tabular}{lcccc}
\toprule
\multirow{2}{*}{\textbf{dataset}} & \multicolumn{2}{c}{\textbf{accuracy$\uparrow$}} & \multicolumn{2}{c}{\textbf{sensitivity$\downarrow$}} \\ \cmidrule{2-5}
                         & Flan-T5-11B           & Flan-T5-770M         & Flan-T5-11B             & Flan-T5-770M          \\ \midrule
CoLA                     & 0.9620        & 0.4307       & 0.0467          & 0.3970        \\
CSQA                     & 0.3090        & 0.7833       & 0.2668          & 0.3516        \\
MNLI                     & 0.7787        & 0.2397       & 0.1659          & 0.4619        \\
RTE                      & 0.6498        & 0.1264       & 0.2366          & 0.4864        \\
SST2                     & 0.9881        & 0.3612       & 0.0203          & 0.3624        \\ \midrule
\textsc{average}         & 0.7375        & 0.3883       & 0.1473          & 0.4119        \\
\bottomrule
\end{tabular}
\caption{The accuracy and sensitivity of Flan-T5-11B and Flan-T5-770M on different tasks with \texttt{zero\_b}. Flan-T5-11B reaches comparable performance to GPT and LLaMA models only on CoLA and SST2.}
\label{t5_zero_b}
\end{table}

\begin{table}[H]
\small
\centering
\begin{tabular}{lccc}
\toprule
\multirow{2}{*}{\textbf{dataset}} & \multirow{2}{*}{\textbf{\#original}} & \multicolumn{2}{c}{\textbf{\#numeric}} \\ \cmidrule{3-4}
                         &                             & Flan-T5-11B   & Flan-T5-770M  \\ \midrule
CoLA                     & 527                         & 496           & 495           \\
CSQA                     & 1220                        & 806           & 1218          \\
MNLI                     & 1000                        & 639           & 924           \\
RTE                      & 277                         & 169           & 208           \\
SST2                     & 872                         & 848           & 497           \\
\bottomrule
\end{tabular}
\caption{The original number of instances in each dataset (\textbf{\#original}) and the number of instances where Flan-T5-11B and Flan-T5-770M output a numeric index using \texttt{base\_a} (\textbf{\#numeric}).}
\label{t5_with_base_a}
\end{table}

\subsection{More on open-ended generation}
\label{more_on_open_ended_generation}

GSM8K is an arithmetic reasoning task in which the outputs are numbers \cite{cobbe2021gsm8k}:

\begin{quote}
\small
\textsc{question:} Sally and Bob have made plans to go on a trip at the end of the year. They both decide to work as babysitters and save half of what they've earned for their trip. If Sally makes \$6 per day and Bob makes \$4 per day, how much money will they both have saved for their trip after a year?
\newline\indent \textsc{answer:} 1825
\end{quote}

\noindent See Table \ref{demonstrations} in \ref{more_on_experiment_settings} for more examples. Table \ref{gsm8k_results} shows the performance of LLaMA2-13B-chat and Flan-T5-11B on GSM8K. There is also a negative correlation between accuracy and sensitivity in open-ended generation. Note that Flan-T5-11B again fails to perform using \texttt{zero\_b}, which is consistent with the results in Section \ref{why_flan_t5_failed_with_zero}.

\begin{table}[h]
\small
\centering
\begin{tabular}{lccc}
\toprule
\textbf{model}                   & \textbf{prompt} & \textbf{accuracy$\uparrow$} & \textbf{sensitivity$\downarrow$} \\ \midrule
\multirow{3}{*}{LLaMA2-13B-chat} & \texttt{base\_b}         & 0.0612            & 0.6226      \\
                                 & \texttt{zero\_b}         & 0.9007            & 0.0015      \\
                                 & \texttt{CoT}             & 0.2570            & 0.6115      \\ \midrule
\multirow{3}{*}{Flan-T5-11B}     & \texttt{base\_b}         & 0.0425            & 0.6635      \\
                                 & \texttt{zero\_b}         & 0.5868            & 0.2802      \\
                                 & \texttt{CoT}             & 0.1208            & 0.6779      \\
\bottomrule
\end{tabular}
\caption{The accuracy and sensitivity of LLaMA2-13B-chat and Flan-T5-11B on GSM8K.}
\label{gsm8k_results}
\end{table}

\subsection{More on gradient-based saliency scores}
\label{more_on_gradient_based_saliency_scores}

\begin{figure}[H]
\centering
\begin{subfigure}{0.48\textwidth}
\includegraphics[width=6cm]{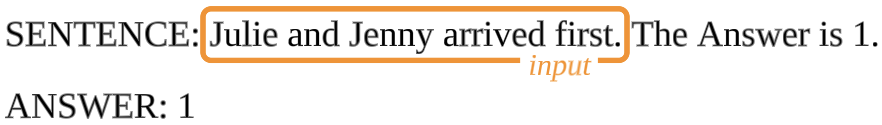}
\caption{\texttt{zero\_b}}
\end{subfigure}
\begin{subfigure}{0.48\textwidth}
\includegraphics[width=7.2cm]{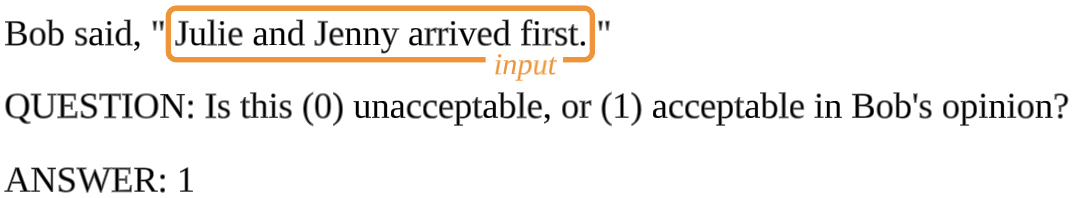}
\caption{\texttt{CFP}}
\end{subfigure}
\caption{Examples of token segmentation in \texttt{zero\_b} and \texttt{CFP}.}
\label{more_on_segmentation}
\end{figure}

\begin{figure}[H]
\begin{subfigure}{\textwidth}
    \includegraphics[width=10cm]{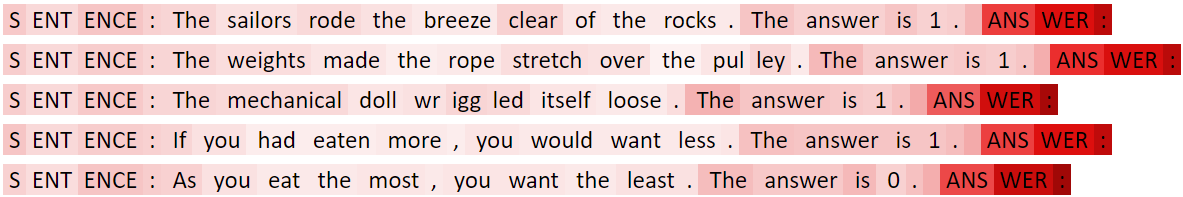}
    \caption{\texttt{zero\_b}}
\end{subfigure}
\begin{subfigure}{\textwidth}
    \includegraphics[width=16cm]{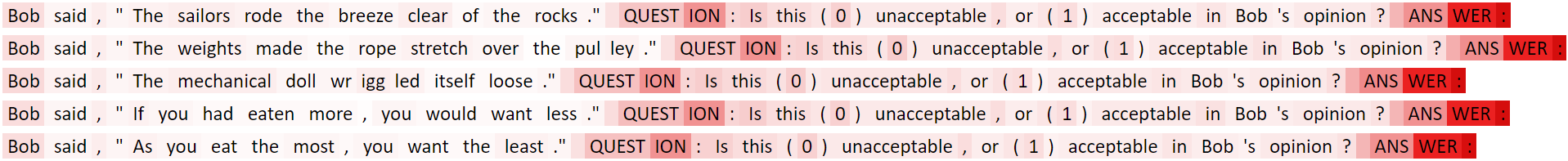}
    \caption{\texttt{CFP}}
\end{subfigure}
\begin{subfigure}{\textwidth}
    \includegraphics[width=16cm]{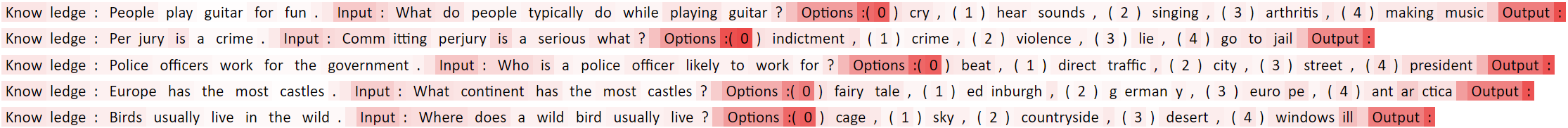}
    \caption{\texttt{GKP}}
\end{subfigure}
\caption{Saliency scores over input tokens of the 5 CoLA and CSQA instances with \texttt{zero\_b}, \texttt{CFP}, and \texttt{GKP} using GPT-6B-JT.}
\label{saliency}
\end{figure}

Figure \ref{more_on_segmentation} show examples of token segmentation for instances with \texttt{zero\_b} and \texttt{CFP}. Figure \ref{saliency} shows the saliency scores over input tokens of CoLA and CSQA instances with \texttt{zero\_b}, \texttt{CFP}, and \texttt{GKP}. Table \ref{mean_saliency_scores_gpt_jt_gpt_j} shows the average \textit{mean saliency scores} of \textit{input} tokens and \textit{prompt} tokens, calculated using GPT-JT-6B, GPT-J-6B, Flan-T5-770M, and T5-770M. Table \ref{mean_saliency_scores_for_target_tokens} shows the average \textit{mean saliency scores} of \textit{input} tokens and ground truth tokens (i.e., tokens in ``The answer is \texttt{[]}.'') of instances with \texttt{zero\_b}.

\begin{table*}[h]
\small
\begin{subtable}[h]{0.5\linewidth}
\centering
\begin{tabular}{lcccc}
\toprule
\multicolumn{1}{l}{\textbf{dataset}} & \multicolumn{1}{c}{\textbf{prompt}}          & $\overline{S}_i$  & $\overline{S}_p$  & $\overline{S}_i / \overline{S}_p$ \\ \midrule
\multirow{3}{*}{CoLA} & \texttt{base\_b} & 4.17 & 12.74 & 32.73
\\
                      & \texttt{zero\_b} & 7.65 & 23.66 & 32.33
\\
                      & \texttt{CFP}     & 3.75 & 11.43 & 32.81
\\ \midrule
\multirow{3}{*}{CSQA} & \texttt{base\_b} & 2.68 & 7.41  & 36.17
\\
                      & \texttt{zero\_b} & 2.25 & 8.02  & 28.05
\\
                      & \texttt{CFP}     & 2.37 & 7.06  & 33.57
\\ \midrule
\multirow{3}{*}{MNLI} & \texttt{base\_b} & 3.10 & 12.79 & 24.24
\\
                      & \texttt{zero\_b} & 3.61 & 17.18 & 21.01
\\
                      & \texttt{CFP}     & 1.68 & 6.71  & 25.04
\\ \midrule
\multirow{3}{*}{RTE}  & \texttt{base\_b} & 1.95 & 9.48  & 20.57
\\
                      & \texttt{zero\_b} & 3.22 & 18.95 & 16.99
\\
                      & \texttt{CFP}     & 1.53 & 7.32  & 20.90
\\ \midrule
\multirow{3}{*}{SST2} & \texttt{base\_b} & 3.09 & 12.59 & 24.54
\\
                      & \texttt{zero\_b} & 4.99 & 22.32 & 22.36
\\
                      & \texttt{CFP}     & 2.77 & 10.70 & 25.89
\\ \midrule
\textsc{average}      & -                & 3.25 & 12.56 & 26.48
\\
\bottomrule
\end{tabular}
\caption{GPT-JT-6B}
\end{subtable}
\begin{subtable}[h]{0.5\linewidth}
\centering
\begin{tabular}{lcccc}
\toprule
\multicolumn{1}{l}{\textbf{dataset}} & \multicolumn{1}{c}{\textbf{prompt}}          & $\overline{S}_i$  & $\overline{S}_p$  & $\overline{S}_i /  \overline{S}_p$ \\ \midrule
\multirow{3}{*}{CoLA} & \texttt{base\_b} & 2.93 & 8.32  & 35.22
\\
                      & \texttt{zero\_b} & 4.08 & 14.34 & 28.45
\\
                      & \texttt{CFP}     & 2.98 & 8.35  & 35.69
\\ \midrule
\multirow{3}{*}{CSQA} & \texttt{base\_b} & 1.82 & 5.01  & 36.33
\\
                      & \texttt{zero\_b} & 1.44 & 4.41  & 32.65
\\
                      & \texttt{CFP}     & 1.91 & 4.64  & 41.16
\\ \midrule
\multirow{3}{*}{MNLI} & \texttt{base\_b} & 2.82 & 9.86  & 28.60
\\
                      & \texttt{zero\_b} & 2.21 & 10.04 & 22.01
\\
                      & \texttt{CFP}     & 1.33 & 5.20  & 25.58
\\ \midrule
\multirow{3}{*}{RTE}  & \texttt{base\_b} & 1.39 & 6.63  & 20.97
\\
                      & \texttt{zero\_b} & 2.01 & 10.94 & 18.37
\\
                      & \texttt{CFP}     & 1.08 & 5.70  & 18.95
\\ \midrule
\multirow{3}{*}{SST2} & \texttt{base\_b} & 2.31 & 8.13  & 28.41
\\
                      & \texttt{zero\_b} & 3.46 & 15.23 & 22.72
\\
                      & \texttt{CFP}     & 2.19 & 6.64  & 32.98
\\ \midrule
\textsc{average}      & -                & 2.26 & 8.23  & 28.54
\\
\bottomrule
\end{tabular}
\caption{GPT-J-6B}
\end{subtable}

\bigskip

\begin{subtable}[h]{0.5\linewidth}
\centering
\begin{tabular}{lcccc}
\toprule
\multicolumn{1}{l}{\textbf{dataset}} & \multicolumn{1}{c}{\textbf{prompt}}  & $\overline{S}_i$  & $\overline{S}_p$  & $\overline{S}_i / \overline{S}_p$ \\ \midrule
CoLA & \texttt{zero\_b} & 7.31 & 21.83 & 33.50
\\
CSQA & \texttt{zero\_b} & 6.03 & 10.27 & 58.69
\\
MNLI & \texttt{zero\_b} & 8.97 & 26.15 & 34.29
\\
RTE  & \texttt{zero\_b} & 6.32 & 24.81 & 25.45
\\
SST2 & \texttt{zero\_b} & 6.71 & 24.04 & 27.91
\\ \midrule
\textsc{average} & -    & 7.07 & 21.42 & 35.97
\\
\bottomrule
\end{tabular}
\caption{Flan-T5-770M}
\end{subtable}
\begin{subtable}[h]{0.5\linewidth}
\centering
\begin{tabular}{lcccc}
\toprule
\multicolumn{1}{l}{\textbf{dataset}} & \multicolumn{1}{c}{\textbf{prompt}}          & $\overline{S}_i$  & $\overline{S}_p$  & $\overline{S}_i / \overline{S}_p$ \\ \midrule
CoLA & \texttt{zero\_b} & 9.56 & 14.52 & 65.84
\\
CSQA & \texttt{zero\_b} & 9.58 & 10.66 & 89.93
\\
MNLI & \texttt{zero\_b} & 8.33 & 11.75 & 70.89
\\
RTE  & \texttt{zero\_b} & 6.49 & 12.94 & 50.17
\\
SST2 & \texttt{zero\_b} & 7.54 & 10.44 & 72.23
\\ \midrule
\textsc{average} & -    & 8.30 & 12.06 & 69.81
\\
\bottomrule
\end{tabular}
\caption{T5-770M}
\end{subtable}
\caption{The average \textit{mean saliency scores} of \textit{input} tokens ($\overline{S}_i$), \textit{prompt} tokens ($\overline{S}_p$), and the ratio between them ($\overline{S}_i / \overline{S}_p$).}
\label{mean_saliency_scores_gpt_jt_gpt_j}
\end{table*}

\begin{table}[H]
\small
\centering
\begin{tabular}{lcccccc}
\toprule
\multirow{2}{*}{\textbf{dataset}} & \multicolumn{2}{c}{$\overline{S}_i$} & \multicolumn{2}{c}{$\overline{S}_t$} & \multicolumn{2}{c}{$\overline{S}_i / \overline{S}_t$} \\ \cmidrule{2-7}
                                  & GPT-JT-6B       & Flan-T5-770M       & GPT-JT-6B       & Flan-T5-770M       & GPT-JT-6B                & Flan-T5-770M               \\ \midrule
CoLA                              & 7.65            & 7.31               & 12.82           & 17.12              & 59.67                    & 42.72                      \\
CSQA                              & 2.25            & 6.03               & 13.50           & 11.60              & 16.67                    & 51.98                      \\
MNLI                              & 3.61            & 8.97               & 13.29           & 23.94              & 27.16                    & 37.46                      \\
RTE                               & 3.22            & 6.32               & 14.63           & 22.74              & 22.01                    & 27.77                      \\
SST2                              & 4.99            & 6.71               & 12.60           & 17.30              & 39.60                    & 38.79                      \\ \midrule
\textsc{average}                  & 3.25            & 7.07               & 13.37           & 18.54              & 24.34                    & 38.12                      \\
\bottomrule
\end{tabular}
\caption{The average \textit{mean saliency scores} of \textit{input} tokens ($\overline{S}_i$), tokens in ``The answer is \texttt{[]}.'' ($\overline{S}_t$), and the ratio between them ($\overline{S}_i / \overline{S}_t$) of instances with \texttt{zero\_b}.}
\label{mean_saliency_scores_for_target_tokens}
\end{table}


\clearpage

\subsection{More on \textit{sensitivity-aware} decoding}
\label{more_on_sensitivity_aware_decoding}

\begin{figure}[h]
\centering
\includegraphics[width=7.2cm]{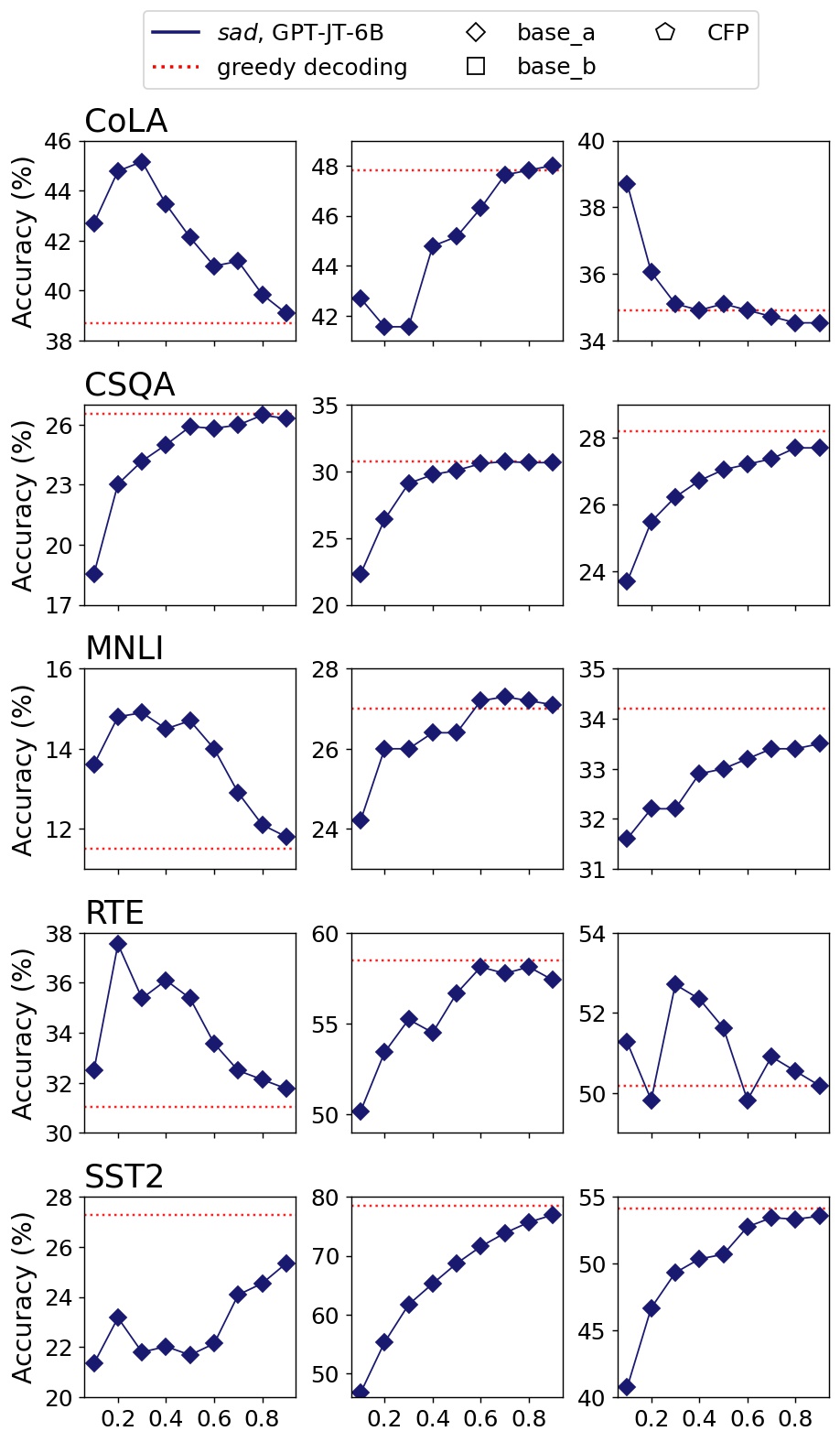}
\caption{Accuracy (\%) of GPT-JT-6B using \texttt{base\_a}, \texttt{base\_b}, and \texttt{CFP} with greedy decoding and \textit{sensitivity-aware} decoding (\texttt{sad}).}
\label{sad_gpt}
\end{figure}

\begin{figure*}[h]
\centering
\includegraphics[width=16cm]{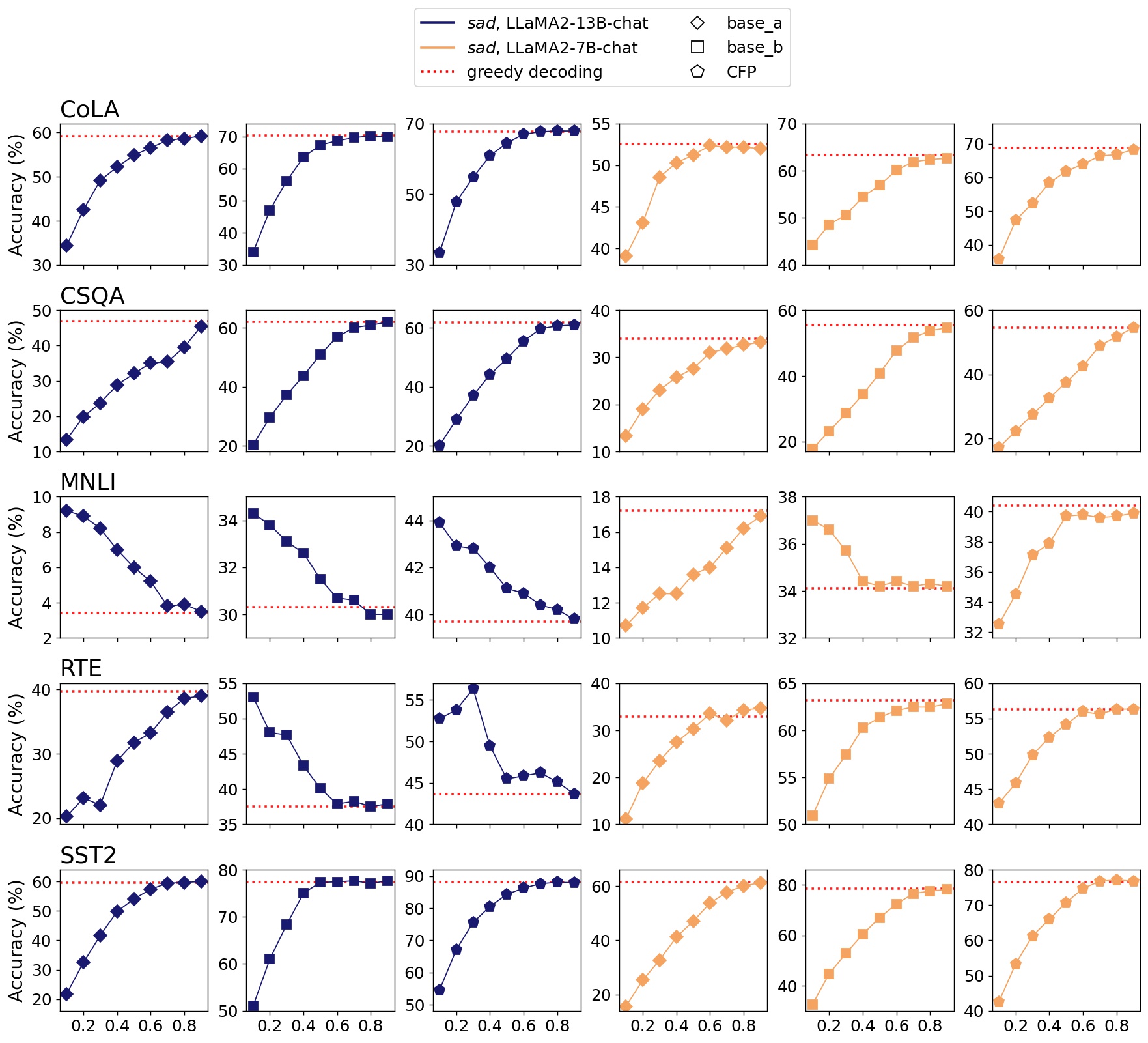}
\caption{Accuracy (\%) of LLaMA2-13B-chat and LLaMA2-7B-chat using \texttt{base\_a}, \texttt{base\_b}, and \texttt{CFP} with greedy decoding and \textit{sensitivity-aware} decoding (\texttt{sad}).}
\label{sad_llama}
\end{figure*}

\begin{figure*}[h]
\centering
\includegraphics[width=16cm]{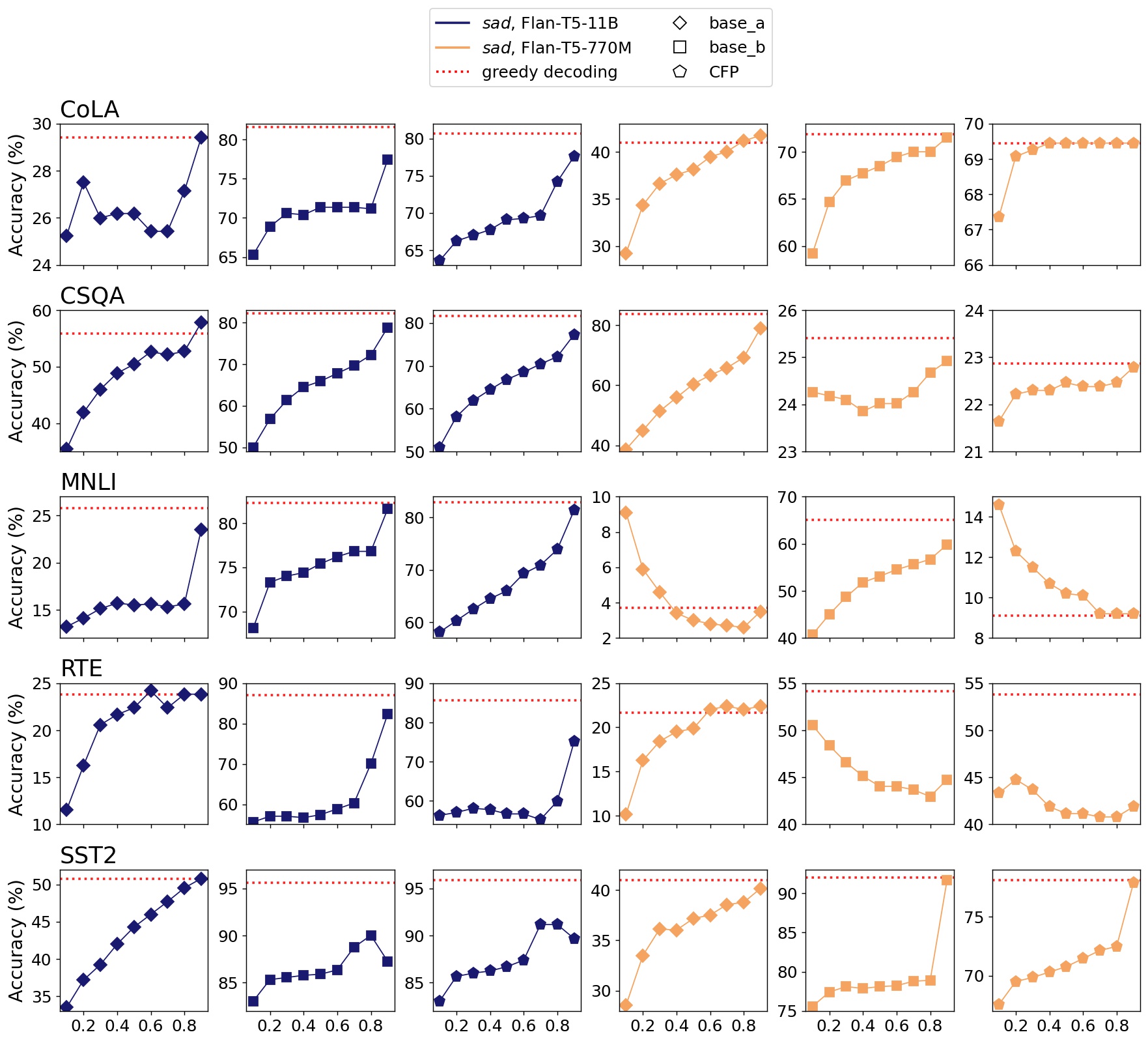}
\caption{Accuracy (\%) of Flan-T5-11B and Flan-T5-770M using \texttt{base\_a}, \texttt{base\_b}, and \texttt{CFP} with greedy decoding and \textit{sensitivity-aware} decoding (\texttt{sad}).}
\label{sad_flan}
\end{figure*}

Figure \ref{sad_gpt}, \ref{sad_llama}, and \ref{sad_flan} show the performance of GPT-JT-6B, LLaMA2-13B-chat, LLaMA2-7B-chat, Flan-T5-11B, and Flan-T5-770M using \texttt{base\_a}, \texttt{base\_b}, and \texttt{CFP} with greedy decoding and \textit{sensitivity-aware} decoding. We experimented with different values of $\alpha$, ranging from $0.1$ to $0.9$. We did five inferences to estimate sensitivity, so the computational costs are five times higher than those of greedy decoding, which only involves a single inference.

\clearpage

\end{document}